\documentclass[journal]{IEEEtran}

\usepackage{amsmath,amssymb,graphicx,paralist,subfigure,pdfpages,cite,algorithm,algpseudocode,paralist,tabularx}
\usepackage{verbatim}
\usepackage{algorithm}
\usepackage{algpseudocode}
\usepackage{cancel}

\newcolumntype{L}[1]{>{\raggedright\arraybackslash}p{#1}}
\newcolumntype{C}[1]{>{\centering\arraybackslash}p{#1}}
\newcolumntype{R}[1]{>{\raggedleft\arraybackslash}p{#1}}

\newtheorem{rmk}{Remark}

\def\diag{{\mbox{diag}}}

\def\bpi{{\boldsymbol \pi}}

\def\ve{\epsilon}
\def\a{\alpha}

\def\ln{{\rm ln}}

\def\mc{\mathcal}
\def\mb{\mathbf}
\def\mbb{\mathbb}
\def\ra{\rightarrow}

\def\AB{\textbf{\texttt{AB}}}
\def\SAB{\textbf{\texttt{SAB}}}
\def\ABm{\textbf{\texttt{ABm}}}
\def\ABN{\textbf{\texttt{ABN}}}
\def\GTS{\textbf{\texttt{GT-SAGA}}}

\IEEEoverridecommandlockouts
\newcommand{\subparagraph}{}
\usepackage[compact]{titlesec}
\titlespacing{\section}{0pt}{2ex}{1ex}
\titlespacing{\subsection}{0pt}{2ex}{0ex}
\titlespacing{\subsubsection}{0pt}{0.5ex}{0ex}

\def\ve{\epsilon}

\def\mbb{\mathbb}
\def\mb{\mathbf}
\def\mc{\mathcal}
\def\wh{\widehat}
\def\wt{\widetilde}
\def\ol{\overline}
\def\ul{\underline}

\def\bds{\boldsymbol}
\newcommand{\mn}[1]{{\left\vert\kern-0.25ex\left\vert\kern-0.25ex\left\vert\kern0.3ex #1 
\kern0.3ex\right\vert\kern-0.25ex\right\vert\kern-0.25ex\right\vert}}

\newcommand{\T}{\top}
\newcommand{\Gra}{\mathcal{G}}

\def\an#1{\color{olive}{#1}}
\def\uk#1{\color{blue}{#1}}

\begin{document}
\title{A general framework for\\  decentralized optimization with first-order methods}
\author{
Ran Xin, Shi Pu, Angelia Nedi\'{c}, and Usman A. Khan
\thanks{
R.\ Xin is with the Electrical and Computer Engineering (ECE) Department at Carnegie Mellon University, Pittsbirgh, PA, USA. 
	S.\ Pu is with School of Data Science, Shenzhen Research Institute of Big Data, The Chinese University of Hong Kong, Shenzhen, China. 
	A.\ Nedi\'{c} is with the School of Electrical, Computer and Energy Engineering at Arizona State University, Tempe, AZ, USA. 
	U.\ A.\ Khan is with the ECE Department at Tufts University, Medford, MA, USA. 
	The work of R.\ Xin and U.\ A.\ Khan has been partially 
	supported by NSF under grants CMMI-1903972 and CBET-1935555. 
	The work of S.\ Pu has been partially supported by the Shenzhen Research Institute of Big Data (SRIBD) Startup Fund No. J00120190011.
	The work by A. Nedi\'c has been supported by NSF grant CCF-1717391.
Email addresses: \texttt{ranx@andrew.cmu.edu, pushi@cuhk.edu.cn, angelia.nedich@asu.edu, khan@ece.tufts.edu.}}}
\maketitle

\begin{abstract}
Decentralized optimization to minimize a finite sum of functions over a network of nodes has been a significant focus within control and signal processing research due to its natural relevance to optimal control and signal estimation problems. More recently, the emergence of sophisticated computing and large-scale data science needs have led to a resurgence of activity in this area. In this article, we discuss decentralized first-order gradient methods, which have found tremendous success in control, signal processing, and machine learning problems, where such methods,
due to their simplicity, serve as the first method of choice for many complex inference and training tasks. In particular, we provide a general framework of decentralized first-order methods that is applicable to undirected and directed communication networks alike, and show that much of the existing work on optimization and consensus can be related explicitly to this framework. We further extend the discussion to decentralized stochastic first-order methods that rely on stochastic gradients at each node and describe how local variance reduction schemes, previously shown to have promise in the centralized settings, are able to improve the performance of decentralized methods when combined with what is known as gradient tracking. We motivate and demonstrate the effectiveness of the corresponding methods in the context of machine learning and signal processing problems that arise in decentralized environments.

\end{abstract}

\section{Introduction}
Minimizing a cost function to select an optimal action or decision has been an important problem in science, engineering, and mathematics. The cost function, say~${F:\mbb R^p\ra\mbb R}$, typically quantifies the loss in fitting data or measurements under a model parameterized by~$\mb x\in\mbb R^p$. An optimal model or decision~$\mb x^*$ is often chosen as the one that minimizes the corresponding loss~$F$. Optimization theory and algorithms~\cite{polyak1987introduction,Nesterov_book,OPTML,OPTbook_LGH} provide the fundamental tools to address such problems. Examples include the classical signal estimation and optimal control problems, where the goal in the former is to minimize the estimation error and in the latter is to minimize the cost of control actions. More recently, with the advent of modern computational machinery, complex nonlinear problems, such as image classification, natural language processing, and deep learning, have enabled a resurgence of interest in the domain of optimization theory and methods. 

Many numerical optimization methods for minimizing a smooth function $F$ are built on a simple observation that moving along the negative gradient~$-\nabla F$ decreases the function $ F$. Thus, given~$\mb x$ and~$\nabla F(\mb x)$, both in~$\mbb{R}^p$, a protocol to decrease~$F(\mb x)$ is~${\mb x_+ = \mb x - \alpha\nabla F(\mb x)}$, if~$\a$, which controls the size of the step taken in the descent direction, is small enough. With the descent in the negative gradient direction, we have that~${F(\mb x_+) \leq F(\mb x)}$. The algorithm that recursively applies the aforementioned protocol is well-known as \textit{gradient descent} and it minimizes the corresponding cost function under certain conditions on the function~$F$ and the step-size~$\a$. Moreover, gradient descent is a \textit{first-order} method as it only uses the first-derivative (gradient) of the cost, in contrast to, for example, second-order methods that typically compute the inverse of the Hessian~$\nabla^2 F$ of the cost function at each iteration.

In this article, we focus on decentralized optimization problems where data samples are available across multiple nodes, such as machines, sensors, robots or mobile devices. The nodes communicate with each other according to a peer-to-peer network, without a central coordinator, and solve the underlying optimization problem in a cooperative manner. Such problems are prevalent in modern-day machine learning where, for example, a large collection of images are stored on multiple machines in a data center for the purpose of image classification. Moreover, classical applications like sensor networks and robotic swarms also fit this paradigm where the sensors and robots collect measurements in order to learn an underlying phenomenon, navigate an environment, or decide on an optimal control action. In such settings, the data samples available at the~$i$th node lead to a local cost~$f_i$, and the goal of the networked nodes is to \textit{agree} on a minimizer of the global cost ${F=\frac{1}{n}\sum_{i=1}^{n}f_i}$ based on the data across all~$n$ nodes. In related applications of practical interest, raw data sharing among the nodes is often not permitted due to the private nature of data, such as text messages or medical images, or is inefficient due to limited communication resources.

Decentralized first-order methods thus rely on information exchange among the nodes and local gradient computation
to build the solution of the global optimization problem. Each node~$i$ iterates on a local state variable~$\mb x_k^i$ that is an estimate of a minimizer~$\mb x^*$ of the global cost function~$F$ at iteration~$k$, and recursively updates this estimate according to the \textit{estimates of neighboring nodes} and \textit{local gradients}. In other words, the nodes do not share their raw data (local gradients) directly and only communicate their estimates (state variables) and perhaps a few other auxiliary variables. Classical solutions along these lines can be found in~\cite{DGD_tsitsiklis,DGD_nedich,DGD_Kar} that are built on average consensus and in~\cite{diffusion_chen} where a diffusion principle is used for agreement. More recent developments include \textit{gradient tracking}~\cite{DAC,GT_CDC,GT_0}, where the local descent direction~$-\nabla f_i$ at each node~$i$ is replaced with a local iterative tracker of the gradient of the global cost function~$\nabla F$. Thus, as these local trackers approach~$\nabla F$, each local iterate~$\mb x_k^i$ descends in the global direction and converges exponentially (linearly on the log scale) to the unique minimizer for smooth and strongly convex problems~\cite{harnessing,DIGing,xin2018linear,push-pull}, in a similar way as the centralized gradient descent. Our primary focus in this article is on the class of smooth and strongly convex problems for the ease of illustrating the key technical ideas. We emphasize however that the decentralized optimization algorithms described herein apply to smooth non-convex problems directly~\cite{NEXT,SS_AB,DSGD_NIPS,D2,improved_DSGT_Xin,GT_SARAH}.

The first half of this article is devoted to decentralized first-order methods based on gradient tracking focusing on a recently introduced algorithm,~$\AB$~\cite{xin2018linear} or \textbf{\texttt{Push-Pull}}~\cite{push-pull}, that utilizes a novel application of both row and column stochastic network weights to achieve linear convergence for smooth and strongly convex problems. Since doubly stochastic weights are not used, the corresponding methods are applicable to both undirected and directed networks. We further describe how $\AB$/\textbf{\texttt{Push-Pull}} unifies much of the existing work on decentralized first-order methods that use gradient tracking and subsumes several non-trivial average consensus algorithms as special cases. Moreover, we emphasize the push and pull communication aspects enabled by the column and row stochastic weights in~$\AB$/\textbf{\texttt{Push-Pull}} and show how $\AB$/\textbf{\texttt{Push-Pull}} unifies various communication architectures. 

The second half of this article is devoted to decentralized stochastic first-order methods, describing the current state-of-the-art and open problems where some progress has been made only recently. In decentralized stochastic gradient methods, each node has access only to an imperfect gradient of its local cost, which results from either an incomplete knowledge of the true gradient or sampling a small subset from a large number of local data samples. In this context, we describe how gradient tracking, previously successful in non-stochastic cases, does not necessarily lead to the same performance improvement and show that exact linear convergence to the global minimum (for smooth and strongly convex problems) can be obtained when gradient tracking is further combined with \textit{variance reduction}, well-known in centralized optimization. We emphasize that much of the existing work on decentralized stochastic gradient methods has focused on undirected networks and the results on directed networks are rather restrictive. 

\subsection{Literature Survey}
Decentralized optimization has been a topic of significant research over the past decade, see e.g.,~\cite{DGD_tsitsiklis,forero2010consensus,xin2018linear,push-pull,DGD_nedich,diffusion_chen,DGD_Kar,6119236,wai2018multi,tutorial_nedich,DOPT_survey_yang}. Methods applicable to undirected networks, based on average consensus protocols~\cite{c_Saber,olshevsky,sam_spl1:15}, include~\cite{DGD_nedich,DGD_Yuan,harnessing,EXTRA} that require doubly stochastic network weight matrices. Relevant work that builds on diffusion principles can be found in~\cite{diffusion_chen,I_chen,II_chen,Exact_Diffusion,Exact_Diffusion_2,NIDS,D2}. 
For arbitrary directed networks, it may not be feasible to construct doubly stochastic weights and hence the corresponding decentralized methods build on consensus with row and/or column stochastic weights. For example, the methods in~\cite{opdirect_Tsianous2,opdirect_Nedic,xi_tac2:17,DIGing,xi_tac3:17,xi_tac4:17,xin_frost:18} need the weights to be column stochastic and use the push-sum correction~\cite{ac_directed0,ac_directed,ac_row} that requires a division with a certain eigenvector estimate of the underlying weight matrix. In contrast, the methods in~\cite{xi_tac1:16,xi_neuro:17} are based on surplus consensus~\cite{ac_Cai1} that employs both row and column stochastic weights simultaneously.

Decentralized stochastic optimization in general can be divided into two types: (i) \textit{online}, where an imprecise (stochastic) gradient is drawn from an underlying probability distribution at each node; or (ii) \textit{offline/batch}, where a finite collection of data samples is available locally at each node and a~stochastic gradient is computed from samples drawn randomly from the local batch. Related work on decentralized online problems can be found, e.g., in~\cite{DSGD_nedich,morral2014success,DGD_Kar,SGP_nedich,DSGD_NIPS,SGP_ICML,I_chen,II_chen,pu2017flocking,D2,SED}. Stochastic optimization over finite data in the centralized settings have garnered a strong research activity where modern methods hinge on certain variance-reduction techniques that leverage the finite sum structure of the cost function to accelerate the standard stochastic gradient descent (\texttt{\textbf{SGD}})~\cite{SGD_nemirovski,SGD_Lan}; see, e.g.,~\cite{SAG,SAGA,SVRG,SARAH,SPIDER,katyusha,S2GD}. Existing variance-reduced decentralized stochastic methods can be found in~\cite{DSA,DAVRG,DSBA,edge_DSA,ADFS,SPM2019_Xin,GTVR_TSP,D_Get,N_DANE,PS_MIQ:2020}.

Although not discussed in this article, second-order methods and algorithms based on the curvature of the cost functions can be found in~\cite{ESOM,Ale_QN,Ale_NN,SUCAG,Async_NN_Wei}. Similarly, ADMM (alternating direction method of multipliers) and other primal-dual methods have also been used in decentralized optimization~\cite{ADMM_Mota,ADMM_Shi,ADMM_Wei,prox_ADMM,PDA_Hong,NIDS,GT_jakovetic,proxPDA_Hong,ADMM_hong}. 
See also related work in~\cite{dual_optimal_uribe,dual_GT,dual_optimal_ICML}, which considers methods based on dual gradients. Methods that incorporate communication and computation imperfection and trade-offs, for example, time-varying and random graphs, asynchronous methods and quantization can be found in~\cite{Dusan_Ne,FS_TVAB:18,Async_DGT,olshevsky2018robust,Ale_quant,Balancing_Wei,async_SGP_0,Async_GP}. Another interesting line of work is to use tools from systems and control theory to analyze decentralized optimization algorithms~\cite{IQC_optimization,GT_system,AB_system,DOPT_system}.

\subsection{Outline of the Article}
We now describe the rest of this article. In Section~\ref{s_pf}, we provide the problem formulation, examples, and preliminaries on convex functions, communication graphs, and nonnegative matrices. Section~\ref{s_dgd} discusses early work on decentralized gradient descent and shows its applicability to directed networks with the help of row and column stochastic weights. We then describe gradient tracking and introduce the~$\AB$/\textbf{\texttt{Push-Pull}} algorithm in Section~\ref{sAB}, which further includes a sketch of the analysis, communication architectures, and accelerated methods. Section~\ref{sABSC} describes how~$\AB$/\textbf{\texttt{Push-Pull}} provides a general framework to capture many first-order methods based on gradient tracking. We then begin the discussion on decentralized stochastic first-order methods over both undirected and directed graphs in Sections~\ref{s_dso} and~\ref{s_dsovr}. We show how gradient tracking alone is unable to ensure exact linear convergence in this setting (Section~\ref{s_dso}), but is subsequently achieved when gradient tracking is further combined with variance reduction (Section~\ref{s_dsovr}). Section~\ref{s_exp} provides a detailed numerical study on the performance, including speed-up, and convergence of related algorithms. Finally, Section~\ref{s_conc} concludes the article.  

The goal of this article is to provide an in-depth overview of decentralized first-order methods and to further expose the reader with rigorous yet intuitive arguments to follow the technical analysis. In several remarks distributed throughout this article, we highlight the analysis techniques, practical aspects, and other salient features of the corresponding algorithms.

\subsection{Notation} 
We use lowercase letters to denote scalars in~$\mbb R$, lowercase bold letters to denote vectors, and uppercase letters to denote matrices. For a vector~${\mb x\in\mbb{R}^p}$, we denote its~$i$th element by~$[\mb x]_i$. For a set~$\mc{S}$, we use~$|\mc{S}|$ to denote its cardinality.
The matrix~$I_p$ is the~${p\times p}$ identity, and~$\mb{1}_p$ (resp.~$\mb 0_p$) is the~$p$-dimensional column vector of all ones (resp. zeros). For two matrices~$X,Y$,~${X\otimes Y}$ denotes their Kronecker product. We use~$\left\|\cdot\right\|_2$ to denote the Euclidean norm of a vector. For two sequences of vectors in~$\mbb{R}^p$, say~$\{\mb{a}_k\}_{k\geq0}$ and~$\{\mb{b}_k\}_{k\geq0}$, we denote~${\mb{a}_k\ra \mb{b}_k}$ or~${\mb{b}_k\ra \mb{a}_k}$ as~${\lim_{k\ra\infty}\|\mb{a}_k-\mb{b}_k\|_2 = 0}$; in particular, we denote~$\mb{a}_k\ra\mb{a}$ as~${\lim_{k\ra\infty}\mb{a}_k = \mb{a}}$ for some $\mb{a}\in\mbb{R}^p$.
The spectral radius of a matrix~$X$ is denoted by~$\rho(X)$, while its spectral norm is denoted by~$\mn{X}_2$. 

\section{Problem Formulation and Motivation}\label{s_pf}
\begin{figure}[!h]
\centering
\includegraphics[width=3.4in]{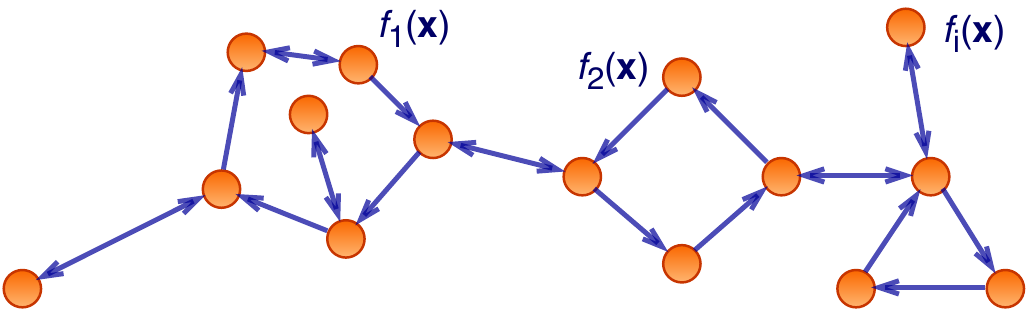}
\caption{Decentralized optimization over a general directed network.}
\label{fig1}
\end{figure}

In this section, we introduce the canonical forms of the decentralized optimization problems and emphasize real-world scenarios where such problems are applicable and essential. Decentralized optimization is finite sum minimization formulated over a network of nodes. Formally, the goal of the nodes is to solve in a cooperative manner 
\begin{align*}
\mbox{Problem P1:}\qquad 
\min_{\mb{x}\in\mbb{R}^p} F(\mb x),\qquad F(\mb x):=\frac{1}{n}\sum_{i=1}^n f_i(\mb x),
\end{align*}
where each~$f_i$ is only locally accessible and processed by node~$i$ and is not shared with any other node. The cooperation (information exchange) among the nodes is peer-to-peer without the existence of a central coordinator and is typically modeled as a graph, see Fig.~\ref{fig1}. In many cases of practical interest, each local cost~$f_i$ can be further decomposed as a weighted sum over local data samples available at node~$i$ and Problem~P1 can be refined as a batch problem:
\begin{align*}
 \mbox{Problem P2:}\qquad 
\min_{\mb{x}\in\mbb{R}^p} &F(\mb x),
\\ F(\mb x)=\frac{1}{n}\sum_{i=1}^n f_i(\mb x)&:=\frac{1}{n}\sum_{i=1}^n
\bigg(\frac{1}{m_i}\sum_{j=1}^{m_i}f_{i,j}(\mb x)\bigg),
\end{align*}
The paradigm of decentralized optimization preserves the privacy of local data and achieves data parallelism, thus enabling effective means for flexible parallel computation. We provide some illustrative examples of this mathematical formulation in the following Section~\ref{s_ex}. For preliminaries on related technical concepts, see Section~\ref{s_pr}.

\vspace{-0.2cm}\subsection{Examples}\label{s_ex}
The finite sum minimization Problems~P1 and~P2 are quite prevalent in signal processing, control, and machine learning. We provide some simple examples below.

\subsubsection{Signal-plus-noise model}
In classical signal processing, we are often interested in finding an unknown signal ${\mb x\in\mbb R^{p}}$ based on the measurements ${y_i=\mb h_i^\top\mb{x}+v_i}$, obtained by a collection of sensors indexed by~$i$, where ${\mb h_i\in\mbb{R}^p}$ is the sensing matrix at sensor~$i$ and~${v_i\in\mbb{R}}$ is measurement noise. Finding~$\mb x\in\mbb{R}^p$ at each sensor~$i$ may be formulated as a local minimization problem in terms of the squared error, i.e.,~$\min_{\mb{x}\in\mbb{R}^p}(y_i-\mb h_i^\top\mb{x})^2$. However, since this problem may be ill-conditioned and the collected measurements have noise, collaboration among the sensors leads to a more robust estimate. The resulting formulation
is
\begin{align*}
\displaystyle\min_{\mb{x}\in\mbb{R}^p}\frac{1}{n}\sum_{i=1}^{n}f_{i}(\mb x),\quad f_{i}(\mb x) := 
(y_i - \mb{h}_i^\top\mb{x})^2,
\end{align*}
which is also known as the least-squares problem.

\subsubsection{Linear models for binary classification}
The decentralized aspects of Problems~P1 and~P2 are more pronounced when we consider high-dimensional and potentially private or proprietary user data. For example, consider a set of~$n$ users that are interested in learning a classifier to distinguish between male and female faces. Each user~$i$~holds a collection of~$m_i$ images, vectorized as~${\mb z_{i,j}\in\mbb{R}^p}$, that are labeled~${y_{i,j}=+1}$ for a male face, or~${y_{i,j}=-1}$ for a female face. The classification may be performed through a linear classifier with parameters~${\mb x\in{\mbb R}^p}$ and~${b\in\mbb R}$, i.e., a hyperplane~$y = \mb{x}^{\top}\mb{z}+b$ that separates the images from two classes. Clearly, a classifier trained on the collection of all images across all users will have superior performance than a locally trained classifier whose performance significantly depends on the size and quality of the local images. 
However, bringing all local images to a central server may be expensive depending on the size of the images and further requires sharing personal information. This discussion motivates the use of decentralized optimization to solve the underlying binary classification problem, for example, with the help of the logistic regression model:
\begin{align*}
\min_{\mb x\in\mbb{R}^p,b\in{\mbb{R}}}~ 
&\frac{1}{n}\sum_{i=1}^n\frac{1}{m_i}\sum_{j=1}^{m_i} \wt f_{i,j}(\mb x) + \frac{\lambda}{2}\|\mb{x}\|_2^2, \nonumber\\
&\wt f_{i,j}(\mb x):=\ln\left[1+\exp\left(-(\mb x^\top\mb z_{i,j}+b)y_{i,j}\right)\right],
\end{align*}
where the logistic loss~$\wt f_{i,j}$ quantifies the error in the linear classifier and~$\frac{\lambda}{2}\|\mb{x}\|^2$, for some $\lambda>0$, is a regularization term to prevent overfitting of the data. 

\subsubsection{Empirical risk minimization}
Problem P2 also arises as an approximation of \textit{expected risk minimization}, see e.g.,~\cite{OPTML} for additional details. In this context, the problem of interest is to find some model~$\mc{H}$, parameterized by~${\mb x\in\mbb{R}^p}$, that maps an input~${\mb z\in\mbb{R}^{{d}_z}}$ to its corresponding output~${\mb y\in\mbb{R}^{{d}_y}}$. The setup requires defining a loss function~${\mc{L}(\mc{H}(\mb{z};\mb x),\mb y)}$ that quantifies the mismatch between the model prediction~$\mc{H}(\mb{z};\mb x)$, under the parameter~$\mb x$, and the actual output data~$\mb{y}$. Assuming that the data~$(\mb z,\mb y)$ belongs to an underlying distribution~$\mc P$, the goal here is to find the optimal parameter~$\mb x^*$ that minimizes the expected loss over~$\mc P$, i.e.,~${\min_{\mb x\in\mbb{R}^p} \mbb E_{(\mb z,\mb y)\sim\mc P}\left[\mc{L}(\mc{H}(\mb z;\mb{x}),\mb y)\right]}$. However, the distribution~$\mc P$ is often intractable in practice and each node~$i$ is \textit{either} able to draw random samples from this distribution in real-time (leading to an online stochastic formulation) \textit{or} has access to a large set of data samples~${(\{\mb z_{i,j}, \mb y_{i,j})\}_{j=1}^{m_i}}$ drawn from~$\mc P$. In the latter case with batch data, the average loss incurred by all data across all nodes serves as an appropriate surrogate for the expected risk and the corresponding problem is often referred to as \textit{empirical risk minimization}, i.e.,
\begin{align*}
\displaystyle\min_{\mb x\in\mbb{R}^p}\frac{1}{n}\sum_{i=1}^{n}\frac{1}{m_i}\sum_{j=1}^{m_i}f_{i,j}(\mb x),\quad f_{i,j}(\mb x) := \mc{L}(\mc{H}(\mb z_{i,j}; \mb x),\mb y_{i,j}).
\end{align*}
This formulation captures a wide range of machine learning models, including deep neural networks.  

\vspace{-0.1cm}
\subsection{Preliminaries}\label{s_pr}
We now briefly describe some mathematical concepts that aid the technical discussion in this article.

\subsubsection{Convex functions}
A convex function~${f:\mbb{R}^p\ra\mbb R}$ is such that for any~$0\!<\!\gamma\!<\!1$ and~$\forall \mb x_1,\mb x_2\in\mbb{R}^p$,
\begin{align}\label{d_cvx}
f(\gamma\mb x_1+(1-\gamma)\mb{x}_2)\leq\gamma f(\mb x_1)+(1-\gamma)f(\mb{x}_2).
\end{align}
The above definition says that a convex function always stays below a line that connects any two points on the function. If~$f$ is differentiable, an equivalent definition of convexity is that it lies above all of its tangents, i.e.,~$\forall \mb x_1,\mb x_2\in\mbb{R}^p$,
\begin{align}\label{d_cvx_tan}
f(\mb x_2) \geq f(\mb x_1) + \nabla f(\mb x_1)^\top(\mb x_2 - \mb x_1),
\end{align}
where~$\nabla f(\mb{x}_1)$ denotes the gradient (derivative) of~$f$ at~$\mb x_1$.
The convexity conditions above are general and do not guarantee the existence of a global minimum. The notion of strong convexity, as defined next, ensures that the global minimum of~$f$ exists and is unique. A function~${f:\mbb{R}^p\ra\mbb R}$ is~$\mu$-strongly convex if~$\forall \mb x_1,\mb x_2\in\mbb{R}^p$,
\begin{align}\label{d_sc}
f(\mb x_2) \geq f(\mb x_1) + \nabla f(\mb x_1)^\top(\mb x_2 - \mb x_1) + \tfrac{\mu}{2}\|\mb x_2-\mb x_1\|_2^2,
\end{align}
for some~${\mu>0}$. It can be verified that strong convexity is stronger than~\eqref{d_cvx_tan} in the sense that it further imposes a quadratic lower bound on~$f$. Finally, a function~${f:\mbb{R}^p\ra\mbb R}$, not necessarily convex, is~$\ell$-smooth if for some~${\ell>0}$ and~${\forall \mb x_1,\mb x_2\in\mbb{R}^p}$,
\begin{align}\label{d_sm1}
\|\nabla f(\mb x_1) - \nabla f(\mb x_2)\|_2 \leq \ell\|\mb x_1 - \mb x_2\|_2,
\end{align}
which implies that 
\begin{align}\label{d_sm2}
f(\mb x_2)\leq f(\mb x_1) + \nabla f(\mb x_1)^\top(\mb x_2 - \mb x_1) + \tfrac{\ell}{2}\|\mb x_2-\mb x_1\|_2^2.
\end{align}
Clearly, an~$\ell$-smooth function has a quadratic upper bound. We denote the class of functions that are both~$\mu$-strongly convex and~$\ell$-smooth as~$\mc S_{\mu,\ell}$. Note that each~${f\in\mc S_{\mu,\ell}}$ is subject to both the lower and upper quadratic bounds in~\eqref{d_sc} and~\eqref{d_sm2}; we thus always have~${\mu\leq \ell}$. The ratio~${\kappa:=\ell/\mu}$ is called the condition number of~$f$ and the functions with large~$\kappa$ are said to be ill-conditioned. 
A rather simple example of a function in this class~$\mc S_{\mu,\ell}$ is~${f(\mb{x})=\mb{x}^\top Q\mb x + \mb{b}^\top\mb{x} + c}$, for~${\mb x\in\mbb R^p}$ and a positive-definite matrix~${Q\in\mbb{R}^{p\times p}}$. The logistic loss~$\wt{f}_{i,j}$ described before is convex and therefore the global cost function becomes strongly-covex with~$\lambda >0$.
See~\cite{Nesterov_book} for more details on convex functions. In this article, the algorithms in question are discussed under the assumption that each local function is smooth and strongly convex, i.e.,
${f_i\in\mc{S}_{\mu,\ell}},{i=1,\ldots,n}$, unless explicitly mentioned otherwise. Thus, the global cost~${F:=\frac{1}{n}\sum_{i=1}^nf_i}$ is such that~${F\in\mc{S}_{\mu,\ell}}$ and we denote the unique minimizer of~$F$ as~$\mb{x}^*$. Where applicable, we refer to the appropriate literature where the convergence results of the related algorithms are generalized to the class of~$\ell$-smooth but possibly non-convex functions.

\subsubsection{Communication Graph}\label{s_gt}
We now formally define the mathematical concept of graphs to characterize the communication (information exchange) among the nodes. Consider~$n$ nodes interacting over a potentially directed graph~${\mc{G}=\{\mc{V,E}\}}$, where~${\mc V := \{1,\ldots,n\}}$ is the set of node indices, and~${\mc E\subseteq \mc V\times \mc V}$ is a collection of ordered pairs~$(i,r)$ such that node~$r$ can send information to node~$i$, i.e.,~${i\leftarrow r}$. Note that~$\mc G$ is not necessarily undirected, i.e.,~${{i\leftarrow r}\nRightarrow {i\ra r}}$. We let~$\mc N_i^{\mbox{\scriptsize in}}$ denote the set of incoming neighbors of node~$i$, i.e., nodes that can send information to node~$i$. Similarly,~$\mc N_i^{\mbox{\scriptsize out}}$ denotes the set of outgoing neighbors, i.e., nodes that receive information from node~$i$. We assume throughout the paper that ${i\in~\mc N_i^{\mbox{\scriptsize in}}\cap\mc N_i^{\mbox{\scriptsize out}}}, \forall i\in\mc{V}$. For an undirected graph, we have that~${\mc N_i:=\mc N_i^{\mbox{\scriptsize in}}=\mc N_i^{\mbox{\scriptsize out}}}$. A directed graph is said to be strongly connected if there exists a directed path between any two nodes.
An undirected graph is said to be connected if it is strongly-connected.
Given a matrix~${{M=\{m_{ir}\}}\in\mathbb{R}^{n\times n}}$, the directed graph induced by the matrix~$M$ is denoted by ${\mathcal{G}_M=\{\{1,2,\ldots,n\},\mathcal{E}_M\}}$, where ${(i,r)\in\mathcal{E}_M}$ if and only if ${m_{ir}>0}$. Conversely, given a directed graph ${\mc{G}=\{\mc{V,E}\}}$, we say~${\ul{W} = \{w_{ir}\}}\in\mbb{R}^{n\times n}$ is a weight matrix associated with~$\mc{G}$ when~${w_{ir}>0}$ if and only if~$(i,r)\in\mc{E}$ and $w_{ir} = 0$ otherwise.

\subsubsection{Nonnegative matrices}
A nonnegative (resp. positive) matrix is such that all of its elements are nonnegative (resp. positive). A matrix is row stochastic (resp. column stochastic) if it is nonnegative and all of its rows (resp. columns) sum to one and it is doubly stochastic if it is both row and column stochastic. The spectral radius of row, column, and doubly stochastic matrices is one,\footnote{This can be shown by, for example, Gershgorin circle theorem~\cite{hornjohnson}.} and one is also an eigenvalue of the corresponding matrix.
A nonnegative matrix~$M$ is irreducible if its induced graph $\mc{G}_M$ is strongly connected and~$M$ is further primitive\footnote{Formally, a nonnegative matrix is said to be primitive if it is irreducible and only has one non-zero eigenvalue of maximum modulus~\cite{hornjohnson}.} if its trace is positive. For a primitive and row stochastic matrix~$\ul A$, from Perron-Frobenius theorem~\cite{hornjohnson}, we denote~$\bpi_A$ as its positive left eigenvector corresponding to the eigenvalue~$1$ such that ${\bpi_A^\top\mb{1}_n = 1}$. Similarly, for a primitive and column stochastic matrix~$\ul B$, we denote~$\bpi_B$ as its positive right eigenvector corresponding to the eigenvalue~$1$ such that~${\mb{1}_n^\top\bpi_B = 1}$. Clearly,
\[
\ul A\mb 1_n=\mb 1_n,~\bpi_A^\top \ul A = \bpi_A^\top, \qquad \ul B\bpi_B=\bpi_B,~\mb 1_n^\top\ul B = \mb 1_n^\top.
\]
and it can be shown that~\cite{hornjohnson}
$$\ul A^\infty:= \lim_{k\ra\infty}\ul{A}^k = \mb{1}_n\bpi_A^\top, \qquad\ul B^\infty := \lim_{k\ra\infty}\ul{B}^k = \bpi_B\mb{1}_n^\top.$$
Additional details on these concepts can be found in~\cite{hornjohnson}.

\section{Decentralized Gradient Descent}\label{s_dgd}
One of the classical methods to minimize a differentiable function~$F:\mbb{R}^p\ra\mbb{R}$ is the gradient descent algorithm \cite{polyak1987introduction}:
\begin{align}\label{cgd}
\mb{x}_{k+1} = \mb x_k - \alpha\cdot\nabla F(\mb x_k), \qquad\forall k\geq 0,   
\end{align}
where~$\{\mb x_k\}_{k\geq0}$ is a sequence of estimates of a minimizer of~$F$ and~${\alpha>0}$ is a constant step-size. When~$F$ is~$\ell$-smooth (but not necessarily convex) and~${\alpha\in(0,\frac{1}{\ell}]}$, it can be shown that ${\|\nabla F(\mb{x}_k)\|_2 \ra 0}$, i.e., gradient descent finds a critical point\footnote{A vector~${\mb x^*\in\mbb R^p}$ is called a critical point of~$F$ if~${\nabla F({\mb x}^*) = \mb 0_p}$.} of~$F$ asymptotically. When~$F$ is further~$\mu$-strongly convex, i.e.,~${F\in\mc{S}_{\mu,\ell}}$, as considered in this article, then a critical point~${\mb x}^*$ is further the unique global minimum of~$F$. In this case, we have that~$\|\mb x_k - \mb{{x}}^*\|_2\leq(1-\mu\a)^k\|\mb x_0-{\mb x}^*\|_2$ for any~$\a\!\in\!(0,\frac{1}{\ell}]$, i.e., gradient descent converges to~$\mb{{x}}^*$ at a linear (on the log-scale) rate~\cite{Nesterov_book,polyak1987introduction}. In the rest of this article, we discuss various decentralized methods of the gradient descent type that are suitable for many practical problems.

When the gradient descent algorithm is implemented locally at node~$i$ without any cooperation to solve the decentralized optimization Problem~P1 and~P2, we note that~\eqref{cgd} only finds the minimizer of~$f_i$ at each node~$i$, but not the minimizer of the global cost~${F=\frac{1}{n}\sum_{i=1}^n f_i}$ in general. A decentralized method thus must have a means of fusing information over the nodes in the network such that the estimate~$\mb x_k^i$, at node~$i$, is steered towards the global minimum~$\mb{x}^*$ of~$F$. In other words, a decentralized optimization algorithm requires two key ingredients: 
\begin{inparaenum}[(i)]
\item \textit{Agreement}: all nodes must agree on the same estimate; and,
\item \textit{Optimality}: the agreement must be on the global minimum.
\end{inparaenum}
Agreement, required by any decentralized optimization algorithm, is typically achieved with the help of
average consensus~\cite{c_Saber,olshevsky,sam_spl1:15} when the underlying communication graph is undirected, or by push-sum~\cite{ac_directed0,ac_directed,ac_row} or surplus consensus~\cite{ac_Cai1} when the underlying graph is directed. We next discuss decentralized optimization methods that build directly on top of these consensus algorithms.

\vspace{-0.2cm}
\subsection{Decentralized gradient descent: Undirected graphs} 
We start with the case of undirected and connected graphs where the construction of doubly stochastic (network) weight matrices in a decentralized manner is straightforward; popular methods include Metropolis and Laplacian weights~\cite{tutorial_nedich}. The most well-known and perhaps the simplest optimization algorithm for solving Problem~P1 and~P2 is Decentralized Gradient Descent (\textbf{\texttt{DGD}})~\cite{DGD_nedich,diffusion_chen,DGD_Kar} described as follows. 
Let ${\mb{x}_k^i\in\mbb{R}^p}$ denote the estimate of~$\mb x^*$ at node~$i$ and iteration~$k$. \textbf{\texttt{DGD}} recursively runs the following iterations at each node~$i$:
\begin{align}\label{dgd}
\mb{x}_{k+1}^i = \sum_{r\in\mc{N}_i}w_{ir}\mb x_k^r - \alpha_k\cdot\nabla f_i(\mb x_k^i), \qquad\forall k\geq 0,  
\end{align}
where~$w_{ir}$ is a weight that node~$i$ assigns to each of its neighboring nodes and the weight matrix~${\ul{W} = \{w_{ir}\}\in\mbb{R}^{n\times n}}$ is doubly stochastic. Note that when~${\a_k=0,\forall k}$, \textbf{\texttt{DGD}} reduces to the classical average consensus, where~${\mb{x}_k^i\ra\frac{1}{n}\sum_{i=1}^n\mb{x}_0^i,\forall i}$, i.e., all nodes agree on the average of their initial states.

To understand \textbf{\texttt{DGD}}, let us consider its vector-matrix form:
\begin{align}\label{dgdv}
\mb{x}_{k+1} = W\mb x_k - \alpha_k \cdot\nabla \mb f(\mb{x}_k), \qquad \forall k\geq0,
\end{align}
where~${W=\ul W\otimes I_p\in\mbb R^{pn\times pn}}$, ${\mb{x}_k\in\mbb R^{pn}}$ concatenates the local $\mb x_k^i$'s, and ${\nabla\mb f(\mb{x}_k)\in\mbb R^{pn}}$ concatenates the local gradients~$\nabla f_i(\mb{x}_k^i)$'s. Assuming that~$\alpha_k\ra0$, it can be shown under certain conditions that $\mb{x}_k^i\ra\overline{\mb x}_k$, where~$\overline{\mb x}_{k}:=\frac{1}{n}\sum_{i=1}^n\mb{x}_k^i$ is the mean iterate over the network. Moreover, multiplying both sides of~\eqref{dgdv} by~$\frac{1}{n}(\mb 1_n^\top\otimes I_p)$, we obtain that
\begin{align}\label{mean_dgd}
\overline{\mb x}_{k+1} = \overline{\mb x}_{k} - \alpha_k\cdot\frac{1}{n}\sum_{i=1}^n\nabla f_i(\mb x_k^i), \qquad \forall k\geq0.
\end{align}
Intuitively we have~${\frac{1}{n}\sum_{i=1}^n\nabla f_i(\mb x_k^i)\ra\nabla F(\overline{\mb x}_{k})}$ as~${\mb{x}_k^i\ra\ol{\mb x}_k}$, at each node~$i$. Therefore, when~${\alpha_k\ra0}$ and the weight matrix~$\ul W$ is doubly stochastic, we have that each~${\mb x_k^i\ra\overline{\mb x}_k}$ and~${\overline{\mb x}_k}$ converges to the minimum of~${F=\frac{1}{n}\sum_{i=1}^n f_i}$ following~\eqref{cgd}, which guarantees simultaneously the agreement and optimality requirement of decentralized optimization. The rigorous analysis of \textbf{\texttt{DGD}} can be found in, e.g.,~\cite{DGD_Yuan,diffusion_chen,tutorial_nedich}. We make a few remarks in the following. 

\begin{rmk}[Performance of \textbf{\texttt{DGD}}--Rate/accuracy tradeoffs]
\textbf{\texttt{DGD}} converges sublinearly to the exact global minimum~$\mb x^*$ of Problem P1 for decaying step-sizes such that~${\alpha_k\ra0}$. Under a constant step-size~$\a$, \textbf{\texttt{DGD}} converges linearly, however, to an inexact solution with accuracy~$\mc{O}(\a)$. It can be shown that a larger constant step-size leads to a faster convergence albeit with worse accuracy; see~\cite{DGD_Yuan,I_chen,II_chen} for details. We will revisit this rate/accuracy tradeoff in Section~\ref{sAB} and show that this issue can be fixed by a technique called gradient tracking.
\end{rmk}

\begin{rmk}[Consensus + innovation and diffusion learning]
\textbf{\texttt{DGD}}-type algorithms of the form~\eqref{dgd} are also known as ``consensus + innovation'' and ``diffusion learning'' in the context of signal estimation problems especially when the cost functions are quadratic; see~\cite{khan_tsp:08,khan_cdc:10,DGD_Kar,ci_kar,diffusion_chen} and references therein. 
\end{rmk}

\vspace{-0.2cm}
\subsection{Decentralized gradient descent: Directed graphs}\label{s_dgd_dg}
We now consider strongly connected, directed graphs where the network weight matrices are either row stochastic or column stochastic but not doubly stochastic, in general.
As a consequence, when the weight matrix in~\eqref{dgd} is column stochastic but not row stochastic, the nodes do not agree since the right eigenvector corresponding to the eigenvalue of~$1$ is not~$\mb{1}_n$ that is essential for agreement or consensus~\cite{xi_tac1:16}. Similarly, when the weights are row stochastic but not column stochastic, the nodes agree however on a sub-optimal solution that is the minimum of a weighted average of local functions (and not the mean). We formally discuss these issues next.

\subsubsection{\textbf{\texttt{DGD}} with column stochastic weights}\label{sDGDCS}
Let~${\ul B=\{b_{ir}\}\in\mbb{R}^{n\times n}}$ be a primitive, column stochastic network weight matrix such that~${\mb 1_n^\top \ul B = \mb 1_n^\top}$ and~${\ul{B}\bpi_{\tiny B}=\bpi_{\tiny B}}$. Consider \textbf{\texttt{DGD}}~\eqref{dgdv} with~$\ul B$, i.e.,
\begin{align}\label{dgdcs}
\mb{x}_{k+1} = B\mb x_k - \alpha_k \cdot\nabla \mb f(\mb{x}_k), \end{align}
where~${B=\ul B\otimes I_p}$. 
Recall that~${\ul{B}^k\ra\ul{B}^\infty:=\bpi_{\tiny B}\mb 1_n^\top}$ and consider the convergence of~\eqref{dgdcs} when~${\a_k = 0,\forall k\geq0}$, i.e.,
\begin{align*}
\mb{x}_{k}\ra B^\infty\mb x_0=(\bpi_{\tiny B}\mb 1_n^\top\otimes I_p)\mb{x}_0&=(\bpi_{\tiny B}\otimes I_p)(\mb 1_n^\top\otimes I_p)\mb{x}_0,
\end{align*}
which shows that~${\mb x_k^i\ra[\bpi_{\tiny B}]_i\sum_{r=1}^n\mb{x}_0^r,\forall i}$. In other words, when the weights are only column stochastic, the nodes do not agree because of the non-identical elements in~$\bpi_{\tiny B}$. We observe however that the scaled iterates~$\mb{x}_k^i/(n[\bpi_{\tiny B}]_i)$ converge to the average of the initial states of the nodes, i.e.,~$\mb{x}_k^i/(n[\bpi_{\tiny B}]_i)\ra \frac{1}{n}\sum_{r=1}^n\mb{x}_0^r,\forall i$. Since no node in the network has the knowledge of the eigenvector~$\bpi_{\tiny B}$, another set of iterations is required to asymptotically estimate~$[\bpi_{\tiny B}]_i$ at each node~$i$. The resulting algorithm is known as \textbf{\texttt{Gradient-Push}}~\cite{opdirect_Tsianous2,opdirect_Nedic}, which runs the following iterations at each node~$i$: 
\begin{subequations}\label{sgp}
\begin{align}\label{sgpa}
z_{k+1}^i &= \sum_{r\in\mc{N}_i^{\tiny\mbox{in}}}b_{ir} z_k^r,\\\label{sgpb}
\mb{x}_{k+1}^i &= \sum_{r\in\mc{N}_i^{\tiny\mbox{in}}}b_{ir}\mb x_k^r - \alpha_k\cdot\nabla f_i(\mb w_{k}^i),\\\label{sgpc}
\mb w_{k+1}^i&=\frac{\mb{x}_{k+1}^i}{z_{k+1}^i},
\end{align}
\end{subequations}
where~${\mb x_0^i = \mb w_0^i\in\mbb R^p}$ is arbitrarily chosen and~${z_0^i=1,\forall i}$. It is straightforward to verify that~${z_k^i\ra n[\bpi_{\tiny B}]_i,\forall i}$, and thus if~${\a_k\ra0}$, intuitively~$\mb{w}_k$ achieves agreement. On the other hand, the column stochasticity of~$\ul{B}$ guarantees the optimality of the \textbf{\texttt{Gradient-Push}} as discussed earlier for~\eqref{mean_dgd}.  

\subsubsection{\textbf{\texttt{DGD}} with row stochastic weights}\label{sDGDRS}
Consider \textbf{\texttt{DGD}} now with row stochastic weights~${\ul{A}=\{a_{ir}\}}$:
\begin{align}\label{dgdrs}
\mb{x}_{k+1} = A\mb x_k - \alpha_k \cdot\nabla \mb f(\mb{x}_k), \end{align}
where~${A=\ul A\otimes I_p}$. We have that~${\ul A \mb 1_n=\mb 1_n}$ and~${\bpi_A^\top \ul A=\bpi_A^\top}$, where~${\bpi_A>0}$ and~${\ul A^k \ra \ul A^\infty:=\mb 1_n\bpi_A^\top}$. In other words, the nodes are able to achieve an agreement under a row stochastic network weight matrix. In particular, without gradient corrections, it is straightforward to show that~${\mb x_{k}^i\ra \sum_{r=1}^n[\bpi_A]_r\mb x_0^r}$, at each node~$i$. In~\eqref{dgdrs}, suppose that~${\alpha_k\ra0}$, we then have that~${\mb x_k^i\ra \sum_{r=1}^n[\bpi_A]_r\mb x_k^r:=\wh{\mb{x}}_k,\forall i}$. Multiplying both sides of~\eqref{dgdrs} by~$(\bpi_A^\top\otimes I_p)$, we obtain that 
\begin{align}\label{xr}
\wh{\mb{x}}_{k+1} = \wh{\mb x}_k - \alpha_k \cdot\sum_{r=1}^n[\bpi_A]_r\nabla f_r(\mb{x}_k^r).
\end{align}
Based on the above discussion, we conclude that each node approaches~$\wh{\mb x}_k$, which converges to the minimum of a \textit{weighted average} of the local costs~$f_i$. The decentralized algorithm to find the minimum of the global cost~$F$ may be constructed by dividing each~$\nabla f_i$ by~$[\bpi_A]_i$. Similar to \textbf{\texttt{Gradient-Push}}, separate iterations however are required to estimate the eigenvector~$\bpi_A$ since it is not locally known at any node. The resulting algorithm~\cite{opdirect_Mai}, termed as~\textbf{\texttt{DGD-RS}}, is given by
\begin{subequations}\label{RSalg}
\begin{align}
\label{dgdrsa}
\mb{x}_{k+1}^i &= \sum_{r\in\mc{N}_i^{\tiny\mbox{in}}}a_{ir}\mb x_k^r - \alpha_k\cdot\frac{\nabla f_i(\mb x_k^i)}{[\mb e_k^i]_i},\\\label{dgdrsb}
\mb e_{k+1}^i &= \sum_{r\in\mc{N}_i^{\tiny\mbox{in}}}a_{ir} \mb e_k^r,
\end{align}
\end{subequations}
where~${\mb x_0^i\in\mbb{R}^p}$ is arbitrary and~${\mb e_0^i\in\mbb R^n}$ is a vector of all zeros except a one in the~$i$th entry. The iterations in~\eqref{dgdrsb} therefore asymptotically estimate the eigenvector~$\bpi_A$ of~$\ul A$. To see that, let~${E_k=[\mb e_k^1~\dots~\mb e_k^ n]^\top}$ with~${E_0=I_n}$ and we note that
\[
E_{k+1} = \ul A E_k \ra \ul A^\infty E_0=\mb 1_n\bpi_A^\top.
\]
Hence, we have~${\mb e_k^i\ra \bpi_A,\forall i}$. However, implementing these iterations requires each node to have a unique identifier in order to select the appropriate element from~$\bpi_A$. 


\newpage
\begin{rmk}[Column and row stochastic weights over directed graphs]
A column stochastic weight matrix~${\ul{B} = \{b_{ir}\}}$ is often constructed as~${b_{ir} = 1/|N_i^{\mbox{\scriptsize out}}|,\forall r\in N_i^{\mbox{\scriptsize out}}}$. This formulation requires each node to know its out degree, i.e.,~$|N_i^{\mbox{\scriptsize out}}|$. A row stochastic weight matrix~${\ul{A}= \{a_{ir}\}}$, on the other hand, can be easily constructed as~${a_{ir} = 1/|N_i^{\mbox{\scriptsize in}}|,\forall r\in N_i^{\mbox{\scriptsize in}}},\forall i$, since each node can locally assign weights to the information it receives. 
\end{rmk}

\begin{rmk}[Average consensus over directed graphs]\label{remps}
It can be easily verified that~\eqref{sgpa}--\eqref{sgpc} recover the average of the initial conditions~$\mb x_0^i$'s when~${\alpha_k=0,\forall k}$. This algorithm is well-known as push-sum that implements average consensus with the help of column stochastic weights~\cite{ac_directed0,ac_directed}.
\end{rmk}

\begin{rmk}[Eigenvector estimation]
Based on the current discussion, we note that over directed graphs there is a certain imbalance that is caused by not having~$\mb 1_n$ as either the left or the right eigenvector corresponding to the eigenvalue~$1$ of the corresponding weight matrices. When the weights are column stochastic, this imbalance manifests itself in disagreement of the estimates, whereas in the row stochastic case, this imbalance causes convergence to the minimum of a weighted sum of local cost functions, in stead of~$F$. To overcome this imbalance, a division by the appropriate eigenvector elements is required that leads to separate iterations for eigenvector estimation. The eigenvector estimation, an iterative procedure in itself, may slow down the convergence of the corresponding algorithms especially when the graphs are not well-connected.
\end{rmk}

\section{The~$\AB$/\textbf{\texttt{Push-Pull}} Framework}\label{sAB}
All three decentralized algorithms discussed in the previous section, with doubly stochastic, column stochastic, and row stochastic weights respectively, converge sub-linearly  to the minimizer~$\mb{x}^*$ with decaying step-sizes, i.e., with~${\alpha_k\ra0}$. With a constant step-size, these methods converge linearly albeit to a sub-optimal solution. The reason for this sub-optimality is that~$\mb x^*$ is not a fixed point when the step-size is a constant, as we explain in the following. Consider~\eqref{dgd} with a constant step-size~$\alpha$ and with~${\mb x_k^i=\mb x^*,\forall i}$ and for some~$k$, then
\begin{align*}
\mb{x}_{k+1}^i = \sum_{r\in\mc{N}_i^{\tiny\mbox{in}}}\!\!w_{ir}\mb x^* - \alpha\cdot\nabla f_i(\mb x^*) = \mb x^* - \alpha\cdot\nabla f_i(\mb x^*)\neq\mb x^*,
\end{align*}
because~${\nabla f_i(\mb x^*)\neq\mb0_p}$ in general, i.e., the minimizer~$\mb{x}^*$ of the global cost function~$F$ does not necessarily minimize the local functions~$f_i's$.
Clearly, this issue disappears if~$\nabla f_i$ at each node is replaced by~$\nabla F$ but that is not possible since~$f_i$'s are distributed and private.

A recently introduced scheme that is referred to as \textit{gradient tracking} overcomes the aforementioned steady-state error, while keeping a constant step-size, by replacing~$\nabla f_i$ with an iterative tracker of the global gradient~$\nabla F$ at each node~$i$~\cite{NEXT,GT_CDC,harnessing,DIGing,xi_tac3:17,MP_scutari}. Formally, a new variable~$\mb y_k^i\in\mbb{R}^p$ is added locally at each node~$i$ to track~${\frac{1}{n}\sum_{i=1}^n \nabla f_i(\mb x_k^i)}$, and the estimate~$\mb{x}_k^i$ of~$\mb{x}^*$ descends in the direction of~$-\mb y_k^i$ in stead of~$\nabla f_i$ as in the \textbf{\texttt{DGD}} framework. The task is thus to design an algorithm that tracks a time-varying function~$\frac{1}{n}\mb \sum_{i=1}^n \nabla f_i(\mb x_k^i)$ whose components are distributed over the network. To this aim, Dynamic Average Consensus~\cite{DAC} is employed that is able to track the sum~$\sum_{i=1}^n \mb{r}_k^i$ of time-varying functions~$\mb r_k^i$, assuming that each~$\mb{r}_k^i$ approaches a constant. 

The resulting algorithm, Decentralized Gradient Descent with Gradient Tracking (\textbf{\texttt{GT-DGD}}), linearly converges to the global minimum~$\mb x^*$ with a constant step-size~$\a$ for smooth and strongly convex problems, i.e., when each~${f_{i}\in\mc S_{\mu,\ell}}$. The update of \textbf{\texttt{GT-DGD}} is given by the following iterations: at each node~$i$ and~$\forall k\geq0$,
\begin{subequations}\label{gtdgd}
\begin{align}\label{gtdgda}
\mb{x}_{k+1}^i &= \sum_{r\in\mc{N}_i^{\tiny\mbox{in}}}w_{ir}\mb x_k^r - \alpha\cdot \mb y_k^i, \\\label{gtdgdb}
\mb y_{k+1}^i &=  \sum_{r\in\mc{N}_i^{\tiny\mbox{in}}}w_{ir}\mb y_k^r + \nabla f_i(\mb x_{k+1}^i) - \nabla f_i(\mb x_{k}^i),
\end{align}
\end{subequations}
where~$\mb{x}_0^i$'s in~$\mbb R^p$ are arbitrary and~${\mb y_0^i = \nabla f_i(\mb x_0^i),\forall i}$. Here, similar to \textbf{\texttt{DGD}}, the network weights~$\{w_{ir}\}$ are chosen such that~${\ul W=\{w_{ir}\}}$ is primitive and doubly stochastic. Intuitively, if~$\mb{x}_k^i \ra \ol{\mb{x}}_k$ at each node~$i$, then~${\sum_{i=1}^n \nabla f_i(\mb x_k^i)}$ approaches the global gradient~$\nabla F(\ol{\mb{x}}_k)$ and we have that~${\mb y_k^i\ra\nabla F(\ol{\mb{x}}_k)}$ at each node~$i$. As a consequence, the iterates obtained via equation~\eqref{gtdgda} asymptotically descends in the global direction~$\nabla F$. However,~\textbf{\texttt{GT-DGD}} is restricted to undirected (or balanced) graphs because the weights~$\ul W$ are required to be doubly stochastic. Recall that the row stochasticity of~$\ul W$ leads to agreement, while the column stochasticity leads to optimality. Although row stochasticity and column stochasticity are both required, we may ask whether they must hold simultaneously for the weight matrices in \eqref{gtdgda} and \eqref{gtdgdb}.

In particular,~\eqref{gtdgda} may be implemented with row stochastic weights (that are not necessarily column stochastic) leading to an agreement among the nodes, while~\eqref{gtdgdb} may be implemented with column stochastic weights (that are not necessarily row stochastic) ensuring optimality. This observation leads to the~$\textbf{\texttt{AB}}$/\textbf{\texttt{Push-Pull}} algorithm\footnote{Algorithm~\ref{ABalg} was proposed parallelly as~$\AB$ in~\cite{xin2018linear} and as \textit{push-pull} in \cite{push-pull}; see Section~\ref{ABpp} for push and pull aspects of the underlying communication.} described formally in Algorithm~\ref{ABalg}, written with the help of row and column stochastic weights,~${\ul A = \{a_{ir}\}}$ and~${\ul B=\{b_{ir}\}}$, respectively. Since the implementation of~$\AB$/\textbf{\texttt{Push-Pull}} does not requite doubly stochastic weights, the algorithm is applicable to both directed and undirected graphs. When each~$f_i\!\in\!\mc S_{\mu,\ell}$, $\AB$/\textbf{\texttt{Push-Pull}} converges linearly to the global minimum~$\mb x^*$ of~$F$~\cite{xin2018linear,push-pull}. 
\begin{algorithm}[!h]
\caption{~$\AB$/\textbf{\texttt{Push-Pull}}: At each node~$i$}
\label{ABalg}
	\begin{algorithmic}[1]
		\Require $\mb x_0^i\in\mbb R^p,\mb{y}_0^i\!=\!\nabla f_i(\mb x_0^i),\alpha>0$.
		\For{$k= 0,1,2,\cdots$}
	    \State \textbf{State update:} $\mb{x}_{k+1}^i = \sum_{r\in\mc{N}_i^{\tiny\mbox{in}}}a_{ir}\mb x_k^r - \alpha\cdot \mb y_k^i$
	    \State \textbf{Gradient tracking update:} 
	    
	    $\!\!\mb y_{k+1}^i =  \sum_{r\in\mc{N}_i^{\tiny\mbox{in}}}b_{ir}\mb y_k^r + \nabla f_i(\mb x_{k+1}^i) - \nabla f_i(\mb x_{k}^i)$
		\EndFor
	\end{algorithmic}
\end{algorithm}

\begin{rmk}[Uncoordinated Step-sizes]
The~$\AB$/\textbf{\texttt{Push-Pull}} algorithm is applicable to the case when each node~$i$ chooses a constant but distinct step-size~$\a_i$. The linear convergence of~$\AB$/\textbf{\texttt{Push-Pull}} can be established when the maximum step-size~$\max_i \a_i$ across all nodes is positive and sufficiently small, while all other step-sizes can in fact be chosen as~$0$; see \cite{pushpull_TAC,ABm,xin_frost:18} for detailed discussion. For the sake of the argument, let us assume that node~$1$ in a network of~$n$ nodes chooses an appropriate step-size~${\a_1>0}$, while~${\a_i=0}$, for~$i=2,\ldots,n$. Then node~$1$ implements a descent in its state update, while all other nodes only implement consensus. The descent direction~$\mb y_k^1$ at node~$1$ however comes from mixing information among all nodes via the gradient tracking update and~$\mb x_k^1$ thus asymptotically descends in the global direction. Because the rest of the nodes  update~$\mb{x}_{k}^i,i=2,\ldots,n$, without the gradient correction, the overall scheme operates in a leader-follower mode, where node~$1$ acts as the leader while the remaining follower nodes converge to the state of the leader node; see~\cite{sam_spl2:15} for details on the leader-follower algorithm.
\end{rmk}

\vspace{-0.2cm}
\subsection{Analysis}\label{s_ABan}
In this section, we briefly describe the main ideas to establish the linear convergence of~$\AB$/\textbf{\texttt{Push-Pull}}. To proceed, we write $\AB$/\textbf{\texttt{Push-Pull}} in a compact form as follows:
\begin{align}\label{ABmva}
\mb{x}_{k+1} &= A\mb{x}_k - \alpha \mb y_k,\\\label{ABmvb}
\mb{y}_{k+1} &= B\mb{y}_k + \nabla \mb f(\mb{x}_{k+1}) - \nabla\mb f(\mb x_k),
\end{align}
where~${A=\ul A \otimes I_p}$,~${B=\ul B\otimes I_p}$, and the vectors $\mb x_k, \mb{y}_k,$ and $\nabla \mb f(\mb x_k)$ concatenate all~$\mb x_k^i$'s,~$\mb{y}_k^i$'s, and~$\nabla f_i(\mb x_k^i)$'s, respectively. Following the arguments of~\textbf{\texttt{DGD}}, we may want to show that~${\mb x_k\ra\mb 1_n\otimes \widehat{\mb{x}}_k}$, where~$\widehat{\mb{x}}_k$ is some weighted network average, and~${\widehat{\mb{x}}_k\ra\mb x^*}$. However, there are two issues with this approach. First,  the weight matrices are not doubly stochastic and hence a contraction in the Euclidean norm (required to establish agreement) is not applicable. In particular,~${\mn{W-W^\infty}_2\!<\!1}$ holds for a primitive doubly stochastic matrix but not necessarily for a row or a column stochastic matrix. Second, the descent is in the direction of~$-\mb y_k$, and~$\mb{x}_k$ is coupled with 
the gradients via~$\mb y_k$.
The formal analysis thus requires an alternate approach and is briefly described next.

\textbf{Step 1--Contracting norms:} Our approach to establish the linear convergence of~$\AB$/\textbf{\texttt{Push-Pull}} is to first find norms in which the row stochastic matrix~$\ul{A}$ and the column stochastic matrix~$\ul{B}$ contract respectively. In this context, note that since both~$\ul A$ and~$\ul B$ are primitive, we use their 
non-$\mb{1}_n$ eigenvectors corresponding to the eigenvalue of~$1$,~$\bds{\pi}_A>0$ and~$\bds{\pi}_B>0$, respectively, to define weighted Euclidean norms as follows:~$\forall \mb{x}\in\mathbb{R}^n$,
\begin{align*}
&\left\|\mb{x}\right\|_{\bds{\pi}_A}\!:=\!\sqrt{[{\bds{\pi}_A}]_1[\mb{x}]_1^2+\dots+[{\bds{\pi}_A}]_n[\mb{x}]_n^2} \!=\! \left\|\mbox{diag}(\sqrt{\bds{\pi}_A})\mb{x}\right\|_2,\\
&\left\|\mb{x}\right\|_{\bds{\pi}_B} :=\sqrt{\tfrac{[\mb{x}]_1^2}{[{\bds{\pi}_B}]_1}+\dots+\tfrac{[\mb{x}]_n^2}{[{\bds{\pi}_B}]_n}} = \left\|\mbox{diag}(\sqrt{\bds{\pi}_B})^{-1}\mb{x}\right\|_2.
\end{align*}
Subsequently, we denote~$\mn{\cdot}_{\bds{\pi}_A}$ and~$\mn{\cdot}_{\bds{\pi}_B}$ as the matrix norms induced by~$\left\|\cdot\right\|_{\bds{\pi}_A}$ and~$\left\|\cdot\right\|_{\bds{\pi}_B}$, respectively~\cite{hornjohnson}, i.e.,
\begin{align*}
&\mn{X}_{\bds{\pi}_A}=\mn{\mbox{diag}(\sqrt{\bds{\pi}_A})X\:\mbox{diag}(\sqrt{\bds{\pi}_A})^{-1}}_2,\\
&\mn{X}_{\bds{\pi}_B}=\mn{\mbox{diag}(\sqrt{\bds{\pi}_B})^{-1}X\:\mbox{diag}(\sqrt{\bds{\pi}_B})}_2,
\end{align*}
$\forall X\in\mathbb{R}^{n\times n}$. With the help of these induced matrix norms, it can be shown that~\cite{khan_cdc1:19}
\begin{align*}
\sigma_A&:=\mn{\ul{A}-\ul{A}^\infty}_{\bds{\pi}_A} < 1,\\ \sigma_B&:=\mn{\ul{B}-\ul{B}^\infty}_{\bds{\pi}_B} < 1.
\end{align*}
See~\cite{khan_cdc1:19} for more details.

\textbf{Step 2--Establish an LTI error dynamics:} With the help of the aforementioned contracting norms for the two weight matrices~$A$ and~$B$, we next describe the three errors embedded in~$\AB$/\textbf{\texttt{Push-Pull}} in their respective norms: 
\begin{enumerate}[(i)]
\item the agreement error~$\|\mb x_k - A^\infty\mb x_k\|_{{\bpi_A}\otimes \mb{1}_p}$; 
\item the optimality gap~$\|A^\infty\mb x_k - \mb 1_n\otimes \mb x^*\|_2$; and,
\item the gradient tracking error~$\|\mb y_k - B^\infty \mb y_k\|_{{\bpi_B}\otimes \mb{1}_p}$.
\end{enumerate}
These errors can be written in a vector
\begin{align*}
\mb t_k = \left[
\begin{array}{c}
\|\mb x_k - A^\infty\mb x_k\|_{\bpi_A\otimes \mb{1}_p}\\
\|A^\infty\mb x_k - \mb 1_n\otimes \mb x^*\|_2\\
\|\mb y_k - B^\infty \mb y_k\|_{\bpi_B\otimes \mb{1}_p}
\end{array}
\right]
\end{align*}
that follows:
\begin{align}\label{tkeq}
\mb t_{k+1} \leq J_\a\mb t_k,
\end{align} where
\begin{align*}
J_\a=\left[
\begin{array}{ccc}
\sigma_A & 0 & 0 \\
0 & 1 & 0\\
a_6 & 0 & \sigma_B
\end{array}
\right]
+\a\left[
\begin{array}{ccc}
a_1 & a_2 &a_3 \\
a_4 & -n\mu(\bpi_A^\top\bpi_B) & a_6\\
a_7& a_8 & a_{9}
\end{array}
\right],
\end{align*}
for some nonnegative constants~$a_i$'s; see~\cite{xin2018linear,push-pull} for details.

\textbf{Step 3--Linear convergence:} Clearly, if~$\mb t_k\ra\mb 0_3$, then~$\mb x_k^i$'s eventually reach agreement and the agreement is on the global minimum~$\mb x^*$ of~$F$, while the rate at which~${\mb t_k\ra \mb 0_3}$ provides an upper bound on the convergence rate of~$\AB$/\textbf{\texttt{Push-Pull}}. Given that~$\mb  t_k$ follows the LTI system in~\eqref{tkeq}, linear convergence of~$\AB$/\textbf{\texttt{Push-Pull}} to~$\mb x^*$ can be established by showing that the spectral radius~$\rho(J_\a)$ of~$J_\a$ is less than one. There are several ways to obtain~${\rho(J_\a)<1}$ for the matrix~$J_\a$ given above and we present the argument from~\cite{xin2018linear} for simplicity; see also~\cite{pushpull_TAC,GTVR_TSP}. First note that when~${\a=0}$,~$J_0$ has three eigenvalues:~$\sigma_A,\sigma_B,1$, two of which are strictly less than~$1$ as given by {Step 1}. Next, since the eigenvalues of a matrix are continuous functions of its elements, it can be shown that as the step-size~$\a$ slightly increases from~$0$, the eigenvalue of~$1$ strictly decreases. In particular, denote by~$q(\a)$ the eigenvalues of~$J_\a$ as a function of~$\a$, we have~that
\[
\left.\frac{d\:q(\a)}{d\a} \right|_{\a=0,q=1} = -n\mu(\bpi_A^\top\bpi_B),
\]
which is strictly negative because~$\bpi_A$ and~$\bpi_B$ are both strictly positive, where~$n$ is the number of nodes and~${\mu>0}$ is the strong convexity constant of~$F$. In other words, we have that~${\rho(J_\a)<1}$ for a sufficiently small~${\a>0}$.

\subsection{Architectures}\label{ABpp}
The~$\AB$/\textbf{\texttt{Push-Pull}} algorithm unifies different types of communication architectures, including decentralized, centralized, and semi-centralized  architecture~\cite{pushpull_TAC,push-pull}. To illustrate, let graphs~$\mathcal{G}_{\ul A}$ and~$\mathcal{G}_{\ul B}$ be induced by the two matrices~${\ul A}$ and~${\ul B}$, respectively, i.e., $a_{ir}>0$ (resp. $b_{ir}>0$) if and only if there exists a link from node $r$ to node $i$ in graph~$\mathcal{G}_{\ul A}$ (reps. $\mathcal{G}_{\ul B}$). Note that when implementing the~$\AB$/\textbf{\texttt{Push-Pull}} algorithm, we have the flexibility to design two different graphs~$\mathcal{G}_{\ul A}$ and~$\mathcal{G}_{\ul B}$, rather than restricting to using one single graph that is commonly considered in the decentralized optimization literature (see e.g. \cite{DIGing}). Such flexibility is key to the unifying property of the~$\AB$/\textbf{\texttt{Push-Pull}} algorithm. 
Formally, we proceed the discussion with the following condition on the graphs~$\mathcal{G}_{\ul A}$ and~$\mathcal{G}_{{\ul B}^\T}$ (the graph~$\mathcal{G}_{{\ul B}^{\T}}$ is the graph~$\mathcal{G}_{{\ul B}}$ with all its edges reversed): the graphs~$\mathcal{G}_{\ul A}$ and~$\mathcal{G}_{{\ul B}^{\T}}$ each contain at least one spanning tree; moreover, there exists at least one node that is a root of spanning trees for both~$\mathcal{G}_{\ul A}$ and~$\mathcal{G}_{{\ul B}^{\T}}$.\footnote{This condition is weaker than assuming both~$\mathcal{G}_{\ul A}$ and~$\mathcal{G}_{\ul B}$ are strongly connected, 
while still imposing certain restrictions on the graph topology. Note that the union of $\mathcal{G}_{\ul A}$ and~$\mathcal{G}_{\ul B}$ is still guaranteed to be strongly connected, which ensures the essential information exchange among the nodes.} 

\begin{figure}[h]
\begin{center}
\includegraphics[height=6.5em,clip = true, trim = 0in 0in 0in 0in]{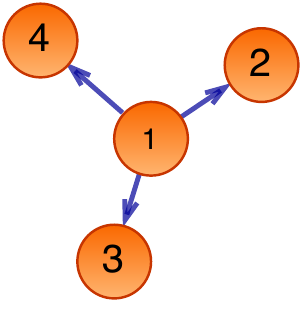}
\qquad\qquad
\includegraphics[height=6.5em,clip = true, trim = 0in 0in 0in 0in]{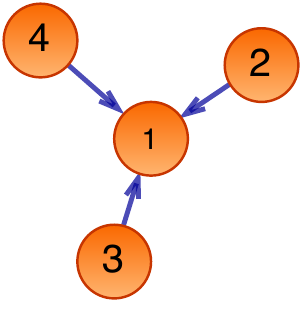}
\caption{On the left is the graph~$\Gra_{\ul A}$ and on the right is the graph~$\Gra_{\ul B}$.}\label{fig:Toy}
\end{center}
\end{figure}

In light of the above conditions on the graphs, we explain how $\AB$/\textbf{\texttt{Push-Pull}} unifies different types of communication architectures using examples from \cite{push-pull}. First, for the fully decentralized architecture, suppose we have a directed and strongly connected graph~$\mathcal{G}$. We can let~$\mathcal{G}_{\ul A}=\mathcal{G}_{{\ul B}}=\mathcal{G}$ and design the weights for~${\ul A}$ and~${\ul B}$ accordingly. Then we have a typical peer-to-peer network structure which is fully decentralized. 
For the case of (semi)-centralized architectures, which is less straightforward, we consider a simple star network $\mathcal{G}$ of four nodes, 
as shown in Fig. \ref{fig:Toy}. Clearly $\Gra_{\ul A}$ and $\mathcal{G}_{{\ul B}^{\T}}$ are identical spanning tree with node $1$ being their common root.
Under this setting, the weight matrices~$\ul A$ and~$\ul{B}$ in~$\AB$/\textbf{\texttt{Push-Pull}} can be chosen as 
\begin{align*}{\ul A}=\left[
{\small \begin{array}{cccc}
1 & 0 &0&0\cr
0.5& 0.5 & 0 & 0\cr
0.5&0&0.5&0\cr
0.5&0&0&0.5
\end{array}}\right],~
{\ul B}=\left[
{\small \begin{array}{cccc}
1&0.5&0.5&0.5\cr
0&0.5&0&0\cr
0&0&0.5&0\cr
0&0&0&0.5
\end{array}}\right].
\end{align*}It can be seen from Fig. \ref{fig:Toy} that the central node~$1$'s information regarding~$\mb x_{k}^1$ is pulled by the entire network through~$\Gra_{\ul A}$; the other nodes only passively receive and use node~$1$'s information. Meanwhile, node~$1$ has been pushed information regarding~$\mb y_{k}^i$ ($i=2,3,4$) from nodes $2$,~$3$, and~$4$ through~$\Gra_{\ul B}$; the other nodes only actively comply with the request from node~$1$. This motivates the algorithm's name {\it \textbf{\texttt{Push-Pull}}} in~\cite{push-pull,pushpull_TAC}. Although nodes~$2$,~$3$, and~$4$ are updating ~$\mb y_k^i$'s accordingly, they do not have to contribute to the optimization process: due to the weights in the last three rows of~${\ul B}$, the values of $\mb y_k^i$'s for nodes~$2$,~$3$, and~$4$ will decrease to $0$ geometrically fast. Without loss of generality, suppose~$f_1(\mb x)=0, \forall \mb x\in\mbb R^p$, and the local stepsize~$\alpha$ for nodes~$2$,~$3$, and~$4$ are set to~$0$. Then the ~$\AB$/\textbf{\texttt{Push-Pull}} algorithm represents a typical centralized gradient method for minimizing~$\sum_{i=2}^4 f_i(x)$ where the master node~$1$ utilizes the slave nodes~$2$,~$3$, and~$4$ to compute the gradient information in a distributed fashion.

Taking the above as a toy example for illustrating the centralized architecture, note that node~$1$ can be replaced by a strongly connected subnet in~$\Gra_{\ul A}$ and~$\Gra_{\ul B}$, respectively. Similarly, all the other nodes can be replaced by subnets as long as the information from the master layer in these subnets can be diffused to all the slave layer nodes in~$\Gra_{\ul A}$, and the information from all the slave layer nodes can be diffused to the master layer in~$\Gra_{\ul B}$. The concept of rooted trees can be used to understand the specific requirements on connectivities of slave subnets. In general, the nodes are referred to as leaders if their roles in the network are similar to the role of node~$1$ in Fig. \ref{fig:Toy}; and the other nodes are referred to as followers. After replacing the individual nodes by subnets, all the subnets have decentralized network structures, while the leader subnet and the follower subnets form a master-slave relationship. This is why such an architecture is called semi-centralized.

\subsection{Acceleration}
It is natural to consider adding momentum to accelerate the convergence of $\AB$/\textbf{\texttt{Push-Pull}}. One immediate extension, based on the Polyak's heavy-ball method~\cite{polyak1987introduction}, is $\ABm$ \cite{ABm} that is given by the following~equations:~$\forall k\geq1$,
\begin{subequations}\label{ABm}
\begin{align}\label{ABma}
\mb{x}_{k+1}^i &= \sum_{r\in\mc{N}_i^{\tiny\mbox{in}}}a_{ir}\mb x_k^r - \alpha\cdot \mb y_k^i + \beta\cdot\mb(\mb{x}_k^i-\mb{x}_{k-1}^i), \\\label{ABmb}
\mb y_{k+1}^i &=  \sum_{r\in\mc{N}_i^{\tiny\mbox{in}}}b_{ir}\mb y_k^r + \nabla f_i(\mb x_{k+1}^i) - \nabla f_i(\mb x_{k}^i),
\end{align}
\end{subequations}
where~${\mb{x}_k^i-\mb{x}_{k-1}^i}$ is the heavy-ball momentum added at each node,~$\alpha$ and~$\beta$ are the step-size and momentum parameters, respectively. It is shown in~\cite{ABm} that~$\ABm$ converges linearly for sufficiently small step-size and momentum parameters, however, acceleration over~$\AB$ is only shown numerically and a detailed theoretical analysis remains an open problem. Alternatively, the use of Nesterov's momentum~\cite{Nesterov_book} is also studied in the context of~$\AB$/\textbf{\texttt{Push-Pull}}. The corresponding algorithm~$\ABN$~\cite{ABN} is given by the following equations:
\begin{subequations}\label{ABN}
\begin{align}
\mb{s}^i_{k+1}&= \sum_{r\in\mc{N}_i^{{\tiny \mbox{in}}}}a_{ir}\mb{x}^{r}_k-\alpha\cdot \mb{y}^i_k, 
\label{ABNa}\\
\mb{x}^i_{k+1}&= \mb{s}^{i}_{k+1} + \beta_k \cdot (\mb{s}^i_{k+1}-\mb{s}^i_{k}), \label{ABNb}\\
\mb{y}^i_{k+1}&= \sum_{r\in\mc{N}_i^{{\tiny \mbox{in}}}}b_{ir}\mb{y}^r_k+\nabla f_i\big(\mb{x}^i_{k+1}\big)-\nabla f_i\big(\mb{x}^i_k\big), \label{ABNd}
\end{align}
\end{subequations}
where~$\beta_k$ is the momentum parameter. When the set of weights above are both doubly stochastic, a closely related form of~$\ABN$ is studied in~\cite{AGT} where acceleration is analytically shown. The analysis of~$\ABN$ however remains an open problem and acceleration is only shown numerically in~\cite{ABN}. 

\section{$\AB$/\textbf{\texttt{Push-Pull}}: A general framework}\label{sABSC}
We now cast~$\AB$/\textbf{\texttt{Push-Pull}} as a general framework that unifies much of the existing work on decentralized first-order methods with gradient tracking. Clearly, \textbf{\texttt{GT-DGD}}~\eqref{gtdgda}--\eqref{gtdgdb} is a special case of~$\AB$/\textbf{\texttt{Push-Pull}} when both sets of weights are further doubly stochastic. We note that \textbf{\texttt{DGD}} also follows from~$\AB$/\textbf{\texttt{Push-Pull}}~\eqref{ABmva}--\eqref{ABmvb}, when~$A$ is replaced by a doubly stochastic matrix~$W$ and~$B$ is chosen as identity~$I_{np}$. Below, we relate~$\AB$/\textbf{\texttt{Push-Pull}} to well-known algorithms in the literature; see~\cite{ABm} for additional discussion. 

\subsection{\textbf{\texttt{GT-DGD}} with column stochastic weights}\label{sGTCS}
We recall~$\AB$/\textbf{\texttt{Push-Pull}} written compactly in~\eqref{ABmva}--\eqref{ABmvb}, where~${A=\ul A\otimes I_p}$ is row stochastic~${B=\ul B\otimes I_p}$ is column stochastic, while both~$\ul A$ and~$\ul B$ are in addition primitive. Define~$\Pi_A$ to be a diagonal matrix with~${\bpi_A\otimes \mb{1}_p}$ on its main diagonal, where~${\bpi_A^\top \ul A=\bpi_A^\top}$ and~${\ul A\mb{1}_n=\mb{1}_n}$. Clearly, we have~${\Pi_A=\diag(\bpi_A)\otimes I_p}$ and note that~${\Pi_A A \Pi_A^{-1}}$ is column stochastic, i.e.,
\begin{align*}
\mb{1}_{np}^\top \cdot \Pi_A A \Pi_A^{-1} &= \mb{1}_{np}^\top (\diag(\bpi_A)\otimes I_p)(\ul A\otimes I_p) \Pi_A^{-1}=\mb{1}_{np}^\top.
\end{align*}
Similarly, we have~${\Pi_A A \Pi_A^{-1} (\bpi_A\otimes\mb{1}_p) = (\bpi_A\otimes\mb{1}_p).}$ With the help of the invertible matrix~$\Pi_A$, we now define a state transformation~$\wt{\mb{x}}_k = \Pi_A\mb x_k$ and modify~\eqref{ABmva}--\eqref{ABmvb} to obtain
\begin{align}\label{BBa}
\wt{\mb{x}}_{k+1} &= \wt{B}\wt{\mb{x}}_k - \alpha\Pi_A \cdot \mb y_k,\\\label{BBb}
\mb{y}_{k+1} &= B\mb{y}_k + \nabla \mb f(\Pi_A^{-1}\wt{\mb{x}}_{k+1}) - \mb f(\Pi_A^{-1}\wt{\mb{x}}_{k}),
\end{align}
where~${\wt{B}:=\Pi_AA\Pi_A^{-1}}$ is column stochastic as we noted earlier. This state transformation shows that a decentralized optimization algorithm with only column stochastic weights~($B$ and~$\wt{B}$) is naturally embedded in~$\AB$/\textbf{\texttt{Push-Pull}}. However, such an algorithm additionally requires the eigenvector vector~$\bpi_A$ of the weights~$\wt B$ in~\eqref{BBa} to implement~\eqref{BBb}. Since~$\bpi_A^\top$ is not locally known to any node, implementing~\eqref{BBa} and~\eqref{BBb} require estimating this non-$\mb 1_n$ right eigenvector of the column stochastic~$\wt B$; not required in~$\AB$/\textbf{\texttt{Push-Pull}} because of the row stochasticity of~$A$. The algorithm that follows is given by, at each node~$i$:
\begin{subequations}\label{Addopt}
\begin{align}\label{Addopta}
\mb{x}_{k+1}^i &= \sum_{r\in\mc{N}_i^{\tiny\mbox{in}}}\wt{b}_{ir}\mb x_k^r - \alpha\cdot \mb y_k^i, \\\label{Addoptb}
z_{k+1}^i &= \sum_{r\in\mc{N}_i^{\tiny\mbox{in}}}\wt{b}_{ir} z_k^r,\\\label{Addoptc}
\mb y_{k+1}^i &=  \sum_{r\in\mc{N}_i^{\tiny\mbox{in}}}b_{ir}\mb y_k^r + \nabla f_i(\tfrac{\mb x_{k+1}^i}{z_{k+1}^i}) - \nabla f_i(\tfrac{\mb x_{k}^i}{z_{k}^i}),
\end{align}
\end{subequations}
where~${\{\wt{b}_{ir}
\}\mbox{ and }\{b_{ir}\}}$ are (possibly different) column stochastic weights. Recall from Section~\ref{s_dgd_dg} that the iterate~$z_k^i$ estimates the right eigenvector~$\bpi_A$ of~${\wt{\ul B}=\{\wt{b}_{ir}\}}$. The above algorithm essentially adds gradient tracking to gradient-push described earlier in~\eqref{sgpa}--\eqref{sgpc}.

\begin{rmk}
The algorithm in~\eqref{Addopta}--\eqref{Addoptc} is well-known as \textbf{\texttt{ADDOPT}}~\cite{xi_tac3:17} and \textbf{\texttt{Push-DIGing}}~\cite{DIGing}, both of which add push-sum consensus to \textbf{\texttt{GT-DGD}} implemented over directed graphs with column stochastic weights. We note that the general idea of push-sum consensus, explored over the \textbf{\texttt{GT-DGD}} framework as in \textbf{\texttt{ADDDOPT/Push-DIGing}}~\cite{xi_tac3:17,DIGing}, essentially results from a state transformation in~$\AB$/\textbf{\texttt{Push-Pull}}. \end{rmk}

\subsection{\textbf{\texttt{GT-DGD}} with row stochastic weights}\label{sGTRS}
Following the previous discussion, we now perform a state transformation on the~$\mb y_k$-update such that the column stochastic weight matrix~$B$ in~$\AB$/\textbf{\texttt{Push-Pull}} transforms into a row stochastic matrix. Let~${\Pi_B := \diag(\bpi_B)\otimes I_p}$, it is easy to verify that~${\wt A:=\Pi_B^{-1}B\Pi_B=\{\wt a_{ir}\}}$ is row stochastic, i.e.,
\begin{align*}
\Pi_B^{-1} B \Pi_B  \mb{1}_{np} &= \mb{1}_{np},\qquad
(\bpi_B^\top\otimes\mb1_p)\Pi_B^{-1} B \Pi_B = \bpi_B^\top\otimes\mb1_p.
\end{align*}
The applicable state transformation~${\wt{\mb y}_k :=\Pi_B^{-1}\mb y_k}$ leads to
\begin{subequations}\label{FROST}
\begin{align}\label{FROSTa}
\mb{x}_{k+1}^i &= \sum_{r\in\mc{N}_i^{\tiny\mbox{in}}}a_{ir}\mb x_k^r - \alpha\cdot \wt{\mb y}_k^i, \\\label{FROSTb}
\mb e_{k+1}^i &= \sum_{r\in\mc{N}_i^{\tiny\mbox{in}}}\wt{a}_{ir} \mb e_k^r,\\\label{FROSTc}
\wt{\mb y}_{k+1}^i &=  \sum_{r\in\mc{N}_i^{\tiny\mbox{in}}}\wt{a}_{ir}\mb y_k^r + \tfrac{\nabla f_i(\mb x_{k+1}^i)}{[\mb e_{k+1}^i]_i} - \tfrac{\nabla f_i(\mb x_{k}^i)}{[\mb e_{k}^i]_i},
\end{align}
\end{subequations}
after adding eigenvector estimation to estimate the ${\mbox{non-}\mb 1_n}$ eigenvector vector of~$\ul B$ corresponding to the eigenvalue~$1$, where~$[\mb e_k^i]_i$ is the~$i$th element of~${\mb e_k^i\in\mbb R^n}$ at node~$i$. The algorithm is initialized with an arbitrary~${\mb x_0^i\in\mbb R^p}$ and~${\mb y_0^i=\nabla f_i(\mb x_0^i)}$, while~${\mb e_0^i}\in\mbb R^n$ is a vector of zeros with a one at the~$i$th location. The resulting algorithm is formally studied in~\cite{xi_tac4:17,xin_frost:18} as \textbf{\texttt{FROST}} and is a gradient tracking extension of \textbf{\texttt{DGD}} with row stochastic weights~\cite{opdirect_Mai}, described earlier in~\eqref{dgdrsa}--\eqref{dgdrsb}.

\subsection{Average consensus}\label{ac_AB}
We now interpret~$\AB$/\textbf{\texttt{Push-Pull}} only for consensus problems and show that it subsumes average consensus over strongly connected graphs as a special case.~To show this, we choose the cost function at each node~$i$ as
\begin{align}\label{avc_fs}
{f}_i(\mb{x})=\tfrac{1}{2}\|\mb{x}-\bds{\upsilon}_i\|_2^2,
\end{align}
for some~$\bds{\upsilon}_i\!\in\!\mbb{R}^p$. Clearly, the minimum of~${F=\sum_{i=1}^{n} f_i}$ is achieved at~${\mb{x}^*=\tfrac{1}{n} \sum_{i=1}^n\bds{\upsilon}_i}$. Thus,~$\AB$/\textbf{\texttt{Push-Pull}} naturally leads to the following average consensus algorithm by noting that~${\nabla\mb{f}(\mb{x}_{k+1})-\nabla\mb{f}(\mb{x}_k)=\mb{x}_{k+1}-\mb{x}_k}$ for the local functions described in~\eqref{avc_fs}: 
\begin{align}\label{sc}
\left[\begin{array}{c}
\mb{x}_{k+1}\\ 
\mb{y}_{k+1}
\end{array}
\right]
= \left[\begin{array}{ccc}
A & -\a I_{np}\\
A - I_{np} & B-\a I_{np}
\end{array}
\right]
\left[\begin{array}{c}
\mb{x}_{k}\\ 
\mb{y}_{k}
\end{array}
\right],
\end{align}
where~${\mb{x}^i_0=\bds\upsilon_i}$ and~${\mb{y}_i^0 = 0,~\forall i}$. From the linear convergence of~$\AB$/\textbf{\texttt{Push-Pull}}, as shown in Section~\ref{s_ABan}, it follows that the above equations converge linearly to the average of~$\bds{\upsilon}_i$'s. What is surprising is that the above form is closely related to a well-known algorithm for average consensus over directed graphs known as \textit{surplus consensus}~\cite{ac_Cai1}. In particular, surplus consensus is obtained after a state transformation of~\eqref{sc} with~$\mbox{diag}\left[I_{np}, -I_{np}\right]$; see~\cite{ABm} for more details.

\begin{rmk}[Consensus over arbitrary graphs]
Following this discussion, choosing the local functions~$f_i$'s as given by~\eqref{avc_fs} in \textbf{\texttt{GT-DGD}}~\cite{GT_CDC,harnessing}, or in \textbf{\texttt{ADDOPT/Push-DIGing}}~\cite{xi_tac3:17,DIGing}, or in \textbf{\texttt{FROST}}~\cite{xi_tac4:17,xin_frost:18}, leads to average consensus, that utilizes gradient tracking, with only doubly stochastic, column stochastic, or row stochastic weights. The protocol that results directly from~$\AB$/\textbf{\texttt{Push-Pull}} is surplus consensus, while the one resulting from \textbf{\texttt{FROST}} has been considered in~\cite{ac_row}. Following the state transformations of Sections~\ref{sGTCS} and~\ref{sGTRS}, it can be verified that the algorithm in~\cite{ac_row} is in fact related to surplus consensus after a state transformation. 
\end{rmk}

\begin{rmk}\label{sc_xi}
Recall that \textbf{\texttt{DGD}}~\eqref{dgd} adds a gradient correction to consensus with doubly stochastic weights and is applicable to undirected graphs. Its extension, ${\mbox{\textbf{\texttt{GP}}~\eqref{sgpa}--\eqref{sgpc}}}$, to directed graphs adds gradient correction to push-sum consensus~\cite{opdirect_Nedic2,opdirect_Nedic}. Similarly, we may use~\eqref{sc}, or equivalently surplus consensus~\cite{ac_Cai1}, as the consensus layer and add a gradient correction to enable decentralized optimization over directed graphs. This idea was explored in~\cite{xi_tac1:16,xi_neuro:17}, however, the convergence rate restrictions are similar to that of \textbf{\texttt{DGD}}.
\end{rmk}

\section{Decentralized Stochastic Optimization}\label{s_dso}
We now discuss stochastic first-order methods in Section~\ref{s_dso} and~\ref{s_dsovr} for Problem~P2. The discussion in Section~\ref{s_dso}, in particular, can be extended to the more general case of online stochastic optimization where a noisy gradient is sampled in real-time; see~\cite{DSGT_Pu} for smooth and strongly convex problems and~\cite{improved_DSGT_Xin} for non-convex problems. 

Recall from Problem~P2 that each node~$i$ possesses~$m_i$ data samples leading to~$m_i$ component costs~$\{f_{i,j}\}_{j=1}^{m_i}$ that constitute the local cost~$f_{i}$. For such problems, the methods described so far in this article are batch~operations, i.e., they compute the gradient~$\nabla f_i$ of the full local cost at each node and thus the descent direction~$\sum_{j_i=1}^{m_i} \nabla f_{i,{j_i}}$ is to be computed over all local data samples~$f_{i,j}$'s. Computing~$m_i$ gradients at each node~$i$ can be taxing particularly when the data is high-dimensional and the cost functions are non-trivial. Computationally-efficient schemes thus rely on stochastic policies to randomly select a small subset of data from the local batch, over which the descent direction is computed. The centralized algorithm based on this idea is the well-known Stochastic Gradient Descent (\textbf{\texttt{SGD}}); see~\cite{OPTML} for a comprehensive review. Assuming that the data is not distributed ($n=1$), and the goal is to minimize~$F=\frac{1}{m}\sum_{j=1}^m f_j$, the centralized \textbf{\texttt{SGD}}, in its simplest form, is given by
\begin{align}\label{sgd}
\mb{x}_{k+1} = \mb x_k - \alpha_k\cdot\nabla f_{\tau_k}(\mb x_k), \qquad\forall k\geq 0,  \end{align}
where~${\tau_k}$ is an index chosen uniformly at random from the set~$\{1,\ldots,m\}$ at each~$k$.

It can be verified that~$\nabla f_{\tau_k}(\mb{x}_k)$ is an unbiased estimate of the full gradient, i.e.,~$\mbb E[\nabla f_{\tau_k}(\mb{x}_k)|\mb{x}_k] = \nabla F(\mb{x}_k)$. Moreover, it is often assumed that each stochastic gradient~$\nabla f_{\tau_k}(\mb{x}_k)$ has uniformly bounded variance, i.e.,
\[
\mbb E\left[\left\|\nabla f_{\tau_k}(\mb x_k)-\nabla F(\mb x_k)\right\|_2^2 | \mb x_k\right] \leq \sigma^2,\qquad {\forall k\ge0},
\]
where~$\sigma>0$ is some universal constant.
Under an additional assumption that~$F\in\mc{S}_{\mu,\ell}$, it can be shown that with a constant step-size~$\alpha\in\left(0,\frac{1}{\ell}\right]$, the mean squared error~$\mathbb{E}\left[\|\mb x_k-\mb x^*\|_2^2\right]$ decays linearly, at the rate of~$\left(1-\mu\alpha\right)^k$, to a neighborhood of~$\mb x^*$. Formally, we have~\cite{OPTML},
\begin{align}\label{sgd_conv}
\mathbb{E}\left[\|\mb x_k - \mb x^*\|_2^2\right]
\leq (1-\mu\alpha)^k + \frac{\alpha{\sigma}^2}{\mu},\qquad \forall k\geq0.
\end{align}
The constant steady-state error given by~$\tfrac{\a\sigma^2}{\mu}$ implies the ``inexact convergence" of \textbf{\texttt{SGD}} due to the persistent gradient noise, i.e., the variance~$\sigma^2$ does not vanish. Diminishing step-sizes overcome this issue and leads to exact convergence albeit at a slower rate. For example, with~$\alpha_k = \frac{1}{\mu (k+1)}$, we have~$\forall k\geq0$,
\begin{align}\label{sgd_conv_2}
\mathbb{E}\left[\|\mb x_k-\mb x^*\|_2^2\right]
\leq \frac{1}{k+1}\max\left\{\frac{2{\sigma}^2}{\mu^2},\|\mb x_0-\mb x^*\|_2^2\right\}.
\end{align}
In other words, to reach an~$\ve$-accurate solution of $\mb x^*$, i.e., ${\mathbb{E}\left[\|\mb x_k-\mb x^*\|^2\right]\leq\ve}$, the \textbf{\texttt{SGD}} method with decaying step-sizes requires~$\mc{O}(\kappa\ve^{-1})$ component gradient evaluations, where~${\kappa = \ell/\mu}$. Note that this rate is independent of the sample size since \textbf{\texttt{SGD}} only computes one component gradient (or a subset of component gradients) per iteration. In the following, we discuss the decentralized versions of \textbf{\texttt{SGD}}. 

\begin{rmk}[\textbf{\texttt{GD}}  vs. \textbf{\texttt{SGD}}]
We note that gradient descent (\textbf{\texttt{GD}})~\eqref{cgd} computes a full gradient, i.e.,~$m$ component gradients per iteration, while \textbf{\texttt{SGD}}~\eqref{sgd} computes only one component gradient (or a subset of component gradients). Consequently, \textbf{\texttt{SGD}} requires~$\mc O(\kappa\ve^{-1})$ component gradient evaluations to reach an~$\ve$-accurate solution of the global minimum~$\mb x^*$, independent of the size~$m$ of the data, versus~$\mc O(m\kappa\:\ln\ve^{-1})$ required by \textbf{\texttt{GD}} that uses the entire batch. It can be argued that \textbf{\texttt{SGD}} is often more preferable in the big data regimes where~$m$ is very large and low-precision solutions may suffice~\cite{OPTML}. Moreover, it is worth noting that \textbf{\texttt{SGD}} is applicable in the more general setting of online, streaming data; see Remark~\ref{sfo_rmk}.
\end{rmk}

\subsection{Decentralized Stochastic Gradient Descent (\textbf{\texttt{DSGD}}):\\ Undirected Graphs}
We now go back to Problem~P1 where the data is distributed over a network of~${n>1}$ nodes and each node~$i$ has access to~$m_i$ component functions indexed by~$j$. Similar to \textbf{\texttt{DGD}}, each node~$i$ samples from its local full batch and implements the following protocol known as \textbf{\texttt{DSGD}}~\cite{DSGD_nedich,diffusion_chen,DGD_Kar}:
\begin{align}\label{dsgd}
\mb{x}^i_{k+1} = \sum_{r\in\mc{N}_i^{\tiny \mbox{in}}}w_{ir}\mb x^r_k - \alpha_k\cdot\nabla f_{i,\tau_k^i}(\mb x_k^i), \qquad\forall k\geq 0,  
\end{align}
where~$\tau_k^i$ is an index chosen uniformly at random from the local index set~$\{1,\ldots,m_i\}$, and the weight matrix~$\ul W=\{w_{ir}\}$ is doubly stochastic. Since the weight matrix is required to be doubly stochastic, the underlying graphs are restricted to be undirected or weight-balanced directed in general. 

As in the centralized case, it is straightforward to verify that $\mbb E\big[\nabla f_{i,\tau_k^i}(\mb{x}_k^i)|\mb{x}_k^i\big] = \nabla f_i(\mb{x}_k^i),\forall i$, and we assume that the variance of each stochastic gradient is uniformly bounded,~i.e.,
\[
\mbb E\Big[\big\|\nabla f_{i,\tau_k^i}(\mb  x_k^i)-\nabla f_i(\mb x_k^i)\big\|_2^2 | \mb x_k^i\Big] \leq \sigma^2,\qquad\forall i,k\geq0.
\]
Assuming that each local function is smooth and strongly convex, i.e.,~$f_{i}\in\mc{S}_{\mu,l},\forall i$, and that~$\lambda$ is the second largest singular value of~$\ul W$, it is shown in~\cite{SED} that under a constant step-size,~$\alpha_k=\alpha\in\big(0,\mc{O}\big(\tfrac{1-\lambda}{\ell\kappa}\big)\big],\forall k$, the mean squared error $\mathbb{E}[\|\mb x_k^i-\mb x^*\|_2^2]$ decays linearly, at the rate of~$\left(1-\mc{O}(\mu\alpha)\right)^k$ up to a steady state error such that
\begin{align}\nonumber
\limsup\limits_{k\ra\infty}&\frac{1}{n}\sum_{i=1}^{n}\mathbb{E}\left[\left\|\mb x_k^i-\mb x^*\right\|_2^2\right]\\\label{DSGD_convergence}
&\qquad\quad=  \mc{O}\left(\frac{\alpha\sigma^2}{n\mu}
+ \frac{\alpha^2\kappa^2\sigma^2}{1-\lambda}
+ \frac{\alpha^2\kappa^2 b}{\left(1-\lambda\right)^2}\right),
\end{align}
where~${b :=\frac{1}{n}\sum_{i=1}^{n}\left\|\nabla f_i\left(\mb x^*\right)\right\|_2^2}$. With a decaying step-size such that~$\alpha_k = \mc{O}(1/k)$, \textbf{\texttt{DSGD}} achieves an exact convergence~\cite{DSGD_Anit,koloskova2019decentralized,pu2019sharp,pu2019asymptotic,olshevsky2018robust}, such that
\begin{align}\label{DSGD_diminishing}
\frac{1}{n}\sum_{i=1}^{n}\mathbb{E}\left[\left\|\mb x_k^i-\mb x^*\right\|_2^2\right] = \mc{O}\left(\frac{1}{k}\right),\qquad\forall k\geq 0.
\end{align} 
See~\cite{pu2019sharp} for explicit non-asymptotic characterization of the mean-squared rate of \textbf{\texttt{DSGD}} under decaying step-sizes and~\cite{DSGD_NIPS} for smooth non-convex problems.

\begin{rmk}[\textbf{\texttt{SGD}} vs. \textbf{\texttt{DSGD}}]\label{brmk}
With decaying step-sizes ${\a_k = \mc{O}(1/k)}$, \textbf{\texttt{DSGD}} achieves a network-independent convergence rate asymptotically that matches the centralized mini-batch \textbf{\texttt{SGD}} (that samples~$n$ stochastic gradients per iteration); the underlying network topology, in fact, only affects the number of transient iterations required to reach this asymptotic rate (see \cite{pu2019sharp,pu2019asymptotic} for precise statements). With an appropriate constant step-size, \textbf{\texttt{DSGD}} achieves linear convergence but to a steady-state error~\eqref{DSGD_convergence}. This steady-state error depends on two additional bias terms when compared with that of centralized \textbf{\texttt{SGD}}~\eqref{sgd_conv}. Both terms arise from the decentralized nature of \textbf{\texttt{DSGD}} and depend on the factor~${(1-\lambda)}$ that encodes the connectivity of the underlying graph. However, the last bias term further includes the constant~$b$ that captures the average deviation between the local and global solutions. In particular, assume~$\mb x_i^*$ to be the locally optimal solution such that~${\nabla f_i(\mb x_i^*)=\mb 0_p},{\forall i}$. We have~${b\leq\tfrac{\ell}{n}\sum_i\|\mb x^*-\mb x_i^*\|^2}$. Clearly,~$b$ is large when data distributions across the nodes are heterogeneous and the local optimal is far from the global optimal. This discussion motivates the use of gradient tracking technique in \textbf{\texttt{DSGD}} as we discuss next. 
\end{rmk}

\subsection{\textbf{\texttt{DSGD}} with Gradient Tracking (\textbf{\texttt{GT-DSGD}}):\\ Undirected Graphs}
As we describe in Section~\ref{sAB}, the performance of \textbf{\texttt{DGD}}~\eqref{dgd} improves with the addition of gradient tracking. Similarly, we can add gradient tracking in \textbf{\texttt{DSGD}} to obtain \textbf{\texttt{GT-DSGD}}, introduced in~\cite{DSGT_Pu,DSGT_MP}, described as follows:
\begin{subequations}
\begin{align}\label{gtdsgda}
\mb{x}^i_{k+1} &= \sum_{r\in\mc{N}_i^{\tiny \mbox{in}}}w_{ir}\mb x^r_k - \alpha_k\cdot\mb{y}_k^i,\\\label{gtdsgdb}
\mb y_{k+1}^i &=  \sum_{r\in\mc{N}_i^{\tiny\mbox{in}}}w_{ir}\mb y_k^r + \nabla f_{i,\tau_{k+1}^i}(\mb x_{k+1}^i) - \nabla f_{i,\tau_k^i}(\mb x_k^i),
\end{align}
\end{subequations}
Similar to \textbf{\texttt{GT-DGD}}, the descent direction~$\mb y_k^i$ at each node~$i$ in \textbf{\texttt{GT-DSGD}} tracks the average of local \textit{stochastic} gradients. In the following, we assume that each~$f_{i,j}$ is~$\ell$-smooth and the global cost~$F$ is~$\mu$-strongly-convex. With a constant step-size $\alpha\in\big(0,\mc{O}\big(\min\big\{1-\lambda,\kappa^{-2/3}\big\}\frac{1-\lambda}{\ell}\big)\big]$, the mean-squared error~$\mathbb{E}\big[\|\mb x_k^i-\mb x^*\|_2^2\big]$ decays linearly, at each node~$i$, at the rate of~$\left(1-\mc{O}(\mu\alpha)\right)^k$ up to a steady-state error such that~\cite{DSGT_Pu}
\begin{equation}\label{DSGT_convergence}
\limsup\limits_{k\ra\infty}\frac{1}{n}\sum_{i=1}^{n}\mathbb{E}\left[\left\|\mb x_k^i-\mb x^*\right\|_2^2\right]
= \mc{O}\left(\frac{\alpha\sigma^2}{n\mu}
+ \frac{\alpha^2\kappa^2\sigma^2}{\left(1-\lambda\right)^3}\right).
\end{equation}
With a decaying step-size such that~$\alpha_k = \mc{O}(1/k)$, we have
\begin{align}\label{DSGT_diminishing}
\frac{1}{n}\sum_{i=1}^{n}\mathbb{E}\left[\left\|\mb x_k^i-\mb x^*\right\|_2^2\right] = \mc{O}\left(\frac{1}{k}\right),\qquad\forall k\geq0.
\end{align} 
See~\cite{improved_DSGT_Xin} for explicit non-asymptotic characterizations and the performance of \textbf{\texttt{GT-DSGD}} for non-convex functions.

\begin{rmk}[\textbf{\texttt{DSGD}} vs. \textbf{\texttt{GT-DSGD}}]\label{dsgdvsgt}
With decaying step-sizes ${\a_k = \mc{O}(1/k)}$, \textbf{\texttt{GT-DSGD}} matches the convergence rate of centralized minibatch \textbf{\texttt{SGD}} after a finite number of transient iterations; see~\cite{improved_DSGT_Xin} for a precise transient time comparison between \textbf{\texttt{GT-DSGD}} and \textbf{\texttt{DSGD}}.
With a constant step-size, the performances of \textbf{\texttt{DSGD}} and \textbf{\texttt{GT-DSGD}} have distinct features. On the one hand, the dependence on~$b$ in the steady-state error of \textbf{\texttt{DSGD}}~\eqref{DSGD_convergence} does not exist in that of \textbf{\texttt{GT-DSGD}}~\eqref{DSGT_convergence}. 
This is intuitive because each local descent direction~$\mb y_k^i$ in \textbf{\texttt{GT-DSGD}} is towards an estimate of the average of local stochastic gradients with the help of gradient tracking; see also Remark~\ref{brmk}.
On the other hand, the steady-state error in \textbf{\texttt{DSGD}} has a better network dependence $\mc O((1-\lambda)^{-2})$ when compared with $\mc O((1-\lambda)^{-3})$ in \textbf{\texttt{GT-DSGD}}. In short, when the data distributions across nodes are substantially heterogeneous, \textbf{\texttt{GT-DSGD}} may be preferable over \textbf{\texttt{DSGD}}.
\end{rmk}

\subsection{Decentralized Stochastic Optimization: Directed Graphs}
When the underlying network is directed, we generally are unable to implement the doubly stochastic weights in \textbf{\texttt{DSGD}} and \textbf{\texttt{GT-DSGD}}. An immediate extension is to use push-sum with \textbf{\texttt{DSGD}} (that uses column stochastic weights) to obtain Stochastic Gradient Push (\textbf{\texttt{SGP}})~\cite{SGP_nedich,SGP_ICML} that replaces the local full batch~$\sum_{j} \nabla f_{i,j}$ in the algorithm described in \eqref{sgpa}--\eqref{sgpc} with a stochastic gradient~\cite{SGP_ICML}. A similar technique can be implemented with only row stochastic weights by writing the algorithm described in~\eqref{dgdrsa}--\eqref{dgdrsb} with stochastic gradients. For the sake of brevity, we only describe the~{\SAB} algorithm here that is a stochastic gradient variant of~\AB/\textbf{\texttt{Push-Pull}}. The basic algorithm is given by
\begin{subequations}\label{SAB}
\begin{align}\label{SABa}
\mb{x}_{k+1}^i &= \sum_{r\in\mc{N}_i^{\tiny\mbox{in}}}a_{ir}\mb x_k^r - \alpha\cdot \mb y_k^i, \\\label{SABb}
\mb y_{k+1}^i &=  \sum_{r\in\mc{N}_i^{\tiny\mbox{in}}}b_{ir}\mb y_k^r + \nabla f_{i,\tau_{k+1}^i}(\mb x_{k+1}^i) - \nabla f_{i,\tau_k^i}(\mb x_{k}^i),
\end{align}
\end{subequations}
where~$\ul A = \{a_{ir}\}$ is row stochastic and~$\ul B=\{b_{ir}\}$ is column stochastic. ${\SAB}$ has been recently introduced in~\cite{khan_cdc1:19}, where it is shown that~$\SAB$ converges linearly with a constant step-size up to a steady-state error. However, a precise characterization of the rate and the steady-state error remains an open problem.

\begin{rmk}[Generalization of~$\SAB$]
Clearly, following the discussion in Section~\ref{sABSC}, one can develop stochastic extensions of \textbf{\texttt{ADDOPT/Push-DIGing}}~\cite{xi_tac3:17,DIGing} and \textbf{\texttt{FROST}}~\cite{xin_frost:18}, where the former only use column stochastic weights (see recent work in~\cite{saddopt} on \textbf{\texttt{S-ADDOPT}} that is a stochastic extension of~\textbf{\texttt{ADDOPT}} for online problems) while the latter only uses row stochastic weights. These extensions rely on eigenvector estimation as described before. Over directed graphs, these variants have the promise of improved performance when compared to the \textbf{\texttt{SGP}}~\cite{SGP_ICML} because of the additional gradient tracking component. However, a formal analysis of~$\SAB$ and the extensions based on row and column stochastic weights remain open problems.
\end{rmk}

\begin{rmk}[Stochastic first-order oracle]\label{sfo_rmk}  
All stochastic gradient methods discussed in this section are applicable to more general \textit{online} stochastic problems, which do not necessarily have the finite sum structure of Problem~P2 on the local costs~$f_i$'s. In particular, each local stochastic gradient in the online case is a noisy sample of the corresponding true gradient~$\nabla f_i(\mb{x}_k^i)$, instead of a randomly chosen component function~$\nabla f_{i,\tau_k^i}(\mb x_k^i)$. Formally, we assume that each node is able to call a Stochastic First-order Oracle (SFO), i.e., at each iteration~$k$ and node~$i$, given~${\mb{x}^i_k\in\mathbb{R}^p}$ as the input, the SFO returns a stochastic gradient~${\mb{g}_i(\mb{x}_{k}^i,\xi_k^i)\in\mathbb{R}^p}$, where~$\xi_k^i$'s are random vectors. The stochastic gradients satisfy the following assumptions:   
The set of random vectors~$\{\xi_k^i\}_{k\geq0,i\in\mc{V}}$ are statistically independent of each other, and 
\begin{enumerate}[(i)]
\item $\mathbb{E}_{\xi_k^i}\left[\mb{g}_i(\mb{x}_{k}^i,\xi_k^i)|\mb{x}_k^i\right]=\nabla f_i(\mb{x}_k^i)$,
\item 
$\mathbb{E}_{\xi_k^i}\left[\left\|\mb{g}_i(\mb{x}_k^i,\xi_k^i)-\nabla f_i(\mb{x}_k^i)\right\|_2^2|\mb{x}_k^i\right]\leq\sigma^2$.
\end{enumerate}
\end{rmk}
See~\cite{SGD_nemirovski,SGD_Lan,OPTML} for additional details and discussion\footnote{The bounded variance condition can be relaxed to (see e.g.,~\cite{SED}) $\mathbb{E}_{\xi_k^i}\left[\left\|\mb{g}_i(\mb{x}_k^i,\xi_k^i)-\nabla f_i(\mb{x}_k^i)\right\|_2^2|\mb{x}_k^i\right]\leq c\|\mb{x}_k^i\|^2+ \sigma^2$, for some~$c > 0$.}. 

\section{Decentralized Stochastic 
First-Order Methods with Variance Reduction}\label{s_dsovr}
The discussion in this section is only applicable to Problem~P2, where the local costs~$f_i$'s have a finite sum structure. In the centralized settings, the corresponding finite sum or batch problem~$\min F(\mb{x})=\tfrac{1}{m}\sum_{j=1}^{m}f_j(\mb{x})$, i.e., Problem~P2 with one node~(${n=1}$), has been a topic of significant research recently. As noted before, \textbf{\texttt{SGD}}~\eqref{sgd} that has served as a promising solution for such problems linearly converges up to a steady-state error~\eqref{sgd_conv} with a constant step-size. This steady-state error is explicitly given by~$\tfrac{\a\sigma^2}{\mu}$ and is a consequence of the variance~$\sigma^2$ of the stochastic gradient. Since the variance is persistent, a plausible way to avoid it is to replace the stochastic gradient in~\eqref{sgd} with an estimator of the full batch gradient. Various variance-reduced methods are developed for the corresponding finite sum batch problems and share a key property that the variance of the gradient estimator goes to zero asymptotically. 

A well-known variance reduction method is \textbf{\texttt{SAGA}}~\cite{SAGA} that replaces the randomly sampled component gradient~$\nabla f_{\tau_k}$ at each iteration of \textbf{\texttt{SGD}}~\eqref{sgd} with an estimator~$\mb g_k$ of the batch gradient. The \textbf{\texttt{SAGA}} method is given by
\begin{subequations}
\begin{align*}
\mb{g}_k &= \nabla f_{\tau_k}(\mb x_k) - \nabla f_{\tau_k}(\wh{\mb x}_{\tau_k}) + \tfrac{1}{m}\textstyle\sum_{j=1}^m\nabla f_j(\wh{\mb x}_{j}),\\
\mb x_{k+1}&=\mb x_k-\alpha\cdot \mb{g}_k,
\end{align*}
\end{subequations}
where~$\wh{\mb x}_j$ is the most recent iterate where the~$j$th component gradient~$\nabla f_j$ was last evaluated and~$\tau_k$ is chosen uniformly at random from the index set~$\{1,\ldots,m\}$, at each iteration~$k$. The implementation of the gradient estimator~$\mb{g}_k$ can be explained as follows. For an arbitrary~${\mb x_0\in\mbb{R}^p}$, compute the component gradients,~$\nabla f_1(\mb x_0), \nabla f_2(\mb x_0), \ldots, \nabla f_m(\mb x_0)$, and store them in a table. At each~${k\geq0}$, draw an index~$\tau_k$, compute~$\mb g_k$ and replace the~$\tau_k$-th element of the gradient table with~$\nabla f_{\tau_k}(\mb x_k)$ while keeping all other entries in the table unchanged. It can be shown that the \textbf{\texttt{SAGA}} estimator is unbiased~\cite{SAGA}, i.e.,~${\mbb E[\mb g_k | \mb x_k] = \nabla F(\mb x_k)}$, and has the property that
\begin{align}\label{VRsaga}
\mbb E\left[\left\|\mb g_k - \nabla F(\mb x_k)\right\|_2^2\right] \ra 0,
\end{align}
as~$\mb x_k\ra\mb{x}^*$. Under the assumption that~${f_j\in\mc{S}_{\mu,\ell},\forall j}$, it can be shown that with~${\alpha = \frac{1}{3\ell}}$, we have~\cite{SAGA},
\begin{align*}
\mathbb{E}\left[\left\|\mb x_k-\mb x^*\right\|_2^2\right]\leq C\left(1-\min\left\{\tfrac{1}{4m},\tfrac{1}{3\kappa}\right\}\right)^k,~~\forall k\geq0,
\end{align*}
for some constant~$C>0$. In other words, \textbf{\texttt{SAGA}} achieves an~$\ve$-accurate solution of~$\mb x^*$ in~$\mc{O}\left(\max\{m,\kappa\}~\ln\ve^{-1}\right)$ component gradient evaluations, where recall that~$\kappa=\frac{\ell}{\mu}$ is the condition number of the global cost~$F$. 

\begin{rmk}[\textbf{\texttt{SGD}} vs \textbf{\texttt{SAGA}}]
\textbf{\texttt{SAGA}}, particularly because of the variance reduction~\eqref{VRsaga}, achieves fast linear convergence to the optimal solution that is independent of the variance of the randomly sampled gradient at each iteration. It is sometimes argued~\cite{OPTML} that when low-precision solutions suffice, \textbf{\texttt{SAGA}}, due to its overhead in terms of gradient storage and relatively slow convergence in the beginning, may not be advantageous enough in comparison with \textbf{\texttt{SGD}} that achieves a fairly decent solution very fast with simpler iterations and no storage needs.
\end{rmk}

\begin{rmk}[Other variance-reduction methods]
Other well-known and popular variance reduction techniques include \textbf{\texttt{SAG}}~\cite{SAG}, \textbf{\texttt{SVRG}}~\cite{SVRG}, \textbf{\texttt{S2GD}}~\cite{S2GD}, \textbf{\texttt{SARAH}}~\cite{SARAH} and \textbf{\texttt{SPIDER}}~\cite{SPIDER}. These methods mostly differ in terms of the gradient estimator and lead to different computation and storage tradeoffs; see~\cite{OPTML} for further discussion. 
\end{rmk}

\subsection{Decentralized Gradient Tracking and Variance Reduction:\\ the \textbf{\texttt{GT-VR}} framework}
Following the discussion so far, it is natural to improve the performance of decentralized stochastic methods with the help of both gradient tracking and variance reduction. Recall that although \textbf{\texttt{GT-DSGD}} is able to eliminate the dependence on~$b$ in the steady state error of \textbf{\texttt{DSGD}}, its linear convergence with a constant step-size is still inexact due to the persistent variance of local stochastic gradients. Hence, the addition of local variance reduction  to~\textbf{\texttt{GT-DSGD}} potentially eliminates this steady-state error. The resulting framework, called \textbf{\texttt{GT-VR}}~\cite{GTVR_TSP}, at each node~$i$ is summarized as:
\begin{enumerate}[(i)]
\item \textit{Local variance reduction}: Draw a local stochastic gradient and update the estimate~$\mb{g}_k^i$ of the local batch gradient $\nabla f_i(\mb{x}_k^i)$ using a variance-reduction procedure;
\item \textit{Global gradient tracking}: Fuse the local batch gradient estimates $\mb g_k^i$'s over nearby nodes to obtain~$\mb{y}_{k+1}^i$ using the gradient tracking protocol;
\item \textit{Local descent}: Update the state~$\mb{x}_{k+1}^i$ as a linear combination of the neighboring states followed by a descent in the direction of~$-\mb{y}_{k+1}^i$.
\end{enumerate}
We can interpret the local variance reduction step as intra-node fusion where each node estimates its own local batch gradient, while the global gradient tracking step as the inter-node fusion where the nodes fuse (the estimate of) their local batch gradients over the network. In essence, the variability in the gradient selection process is removed by performing these two fusions and the overall descent is in the direction of the global negative gradient~$-\nabla F$, asymptotically. 

We next describe an instance of the~\textbf{\texttt{GT-VR}} framework, called \textbf{\texttt{GT-SAGA}}~\cite{GTVR_TSP} that uses \textbf{\texttt{SAGA}}-type local variance reduction. Each node~$i$ starts with an arbitrary~${\mb x_0^i\in\mbb R^p}$, maintains a local gradient table $\big[\nabla f_{i,1}(\mb{x}_0^i),~\ldots~,\nabla f_{i,m_i}(\mb{x}_0^i)\big]$, and computes~${\mb y_0^i = \nabla f_{i}(\mb x_0^i)}$. Node~$i$ then iteratively performs the following steps at each~$k\geq 0$:
\begin{enumerate}[(i)]
\item Perform the~$\mb x_k$-update:
\begin{align}\label{gtsagaa}
\mb x_{k+1}^i = \sum_{r\in\mc N_i}w_{ir}\mb x_{k}^r - \alpha\mb{y}_{k}^{i};
\end{align}
\item Draw~$\tau_k^i$ uniformly at random from~$\{1,\ldots,m_i\}$;
\item Local variance reduction:
\begin{align}\label{gtsagab}
\hspace{-1cm}\mb{g}_{k+1}^i = \nabla f_{i,\tau_{k}^i}\big(\mb x_{k+1}^{i}\big) - \nabla \wh{f}_{i,\tau_{k}^i} + \tfrac{1}{m_i}\textstyle\sum_{j=1}^{m_i}\nabla \wh{f}_{i,j},
\end{align}
where~$\wh{f}_{i,\tau_{k}^i}$ is the~$\tau_k^i$-th element in the gradient table;
\item Global gradient tracking: 
\begin{align}\label{gtsagac}
\mb{y}_{k+1}^{i} &= \sum_{r\in\mc N_i}b_{ir}\mb{y}_{k}^{r} + \mb{g}_{k+1}^i - \mb{g}_k^{i};
\end{align}
\item Replace~$\wh{f}_{i,\tau_k^i}$ with~$\nabla f_{i,\tau_k^i}(\mb x_{k+1}^i)$ in the gradient table.
\end{enumerate}
Similar to the centralized \textbf{\texttt{SAGA}}~\cite{SAGA},~$\GTS$ converges linearly to~$\mb x^*$ with a constant step-size. Formally, when each~$f_{i,j}$ is~$\ell$-smooth and the global cost~$F$ is~$\mu$-strongly-convex, and the step-size is chosen such that~${\alpha = \mc{O}\left(\min\left\{\tfrac{1}{\mu M},\tfrac{m}{M}\frac{(1-\lambda)^2}{\kappa \ell}\right\}\right)}$, we have that~${\forall k\geq 0}$,
\begin{align}
\frac{1}{n}&\sum_{i=1}^{n}\mathbb{E}\left[\left\|\mb x_k^i-\mb x^*\right\|_2^2\right]\nonumber\\ &\leq R\left(1-\min\left\{\mc{O}\left(\tfrac{1}{M}\right),\mc{O}\left(\tfrac{m}{M}\tfrac{(1-\lambda)^2}{\kappa^2}\right)\right\}\right)^k,
\end{align}
where~$m=\min_i\{m_i\},M=\max_i\{m_i\}$, and~$R> 0$ is some constant~\cite{GTVR_TSP}. In other words,~$\GTS$ achieves an~$\ve$-accurate solution of~$\mb x^*$ in 
\begin{align}\label{GTSAGA_comp}
\mc{O}\left(\max\left\{M,\tfrac{M}{m}\tfrac{\kappa^2}{(1-\lambda)^2}\right\}\ln\ve^{-1}\right)
\end{align}
component gradient computations at each node.

\begin{rmk}[Non-asymptotic linear speedup of \textbf{\texttt{GT-SAGA}}] 
When each node has a large data set such that ${M \!\approx\! m \!\geq\! \frac{\kappa^2}{1-\lambda^2}}$, the complexity~\eqref{GTSAGA_comp} of \textbf{\texttt{GT-SAGA}} becomes~${\mc{O}(M\:\ln{\ve^{-1}})}$, independent of the network, and is further~$n$ times faster than that of the centralized \textbf{\texttt{SAGA}} algorithm where it is~${\mc O(nM\:\ln{\ve^{-1}})}$~\cite{SAGA}. Clearly, in this ``big-data" regime, \textbf{\texttt{GT-SAGA}} acts effectively as a means for parallel computation and achieves linear speed-up compared with centralized variance-reduced methods. We also numerically show the linear speedups of decentralized algorithms in Section~\ref{s_exp}. We finally note that other instances of the \textbf{\texttt{GT-VR}} framework also achieve non-asymptotic linear speedup in certain regimes; see~\cite{SPM2019_Xin,GT_SARAH,GTVR_TSP} for details.
\end{rmk}

\begin{rmk}[\textbf{\texttt{DSGD}}, \textbf{\texttt{GT-DSGD}}, and \GTS]
Recapping the performances of the decentralized stochastic first-order methods, we note that~{\GTS} achieves linear convergence to~$\mb{x}^*$ under a constant step-size without a steady-state error due to the local variance-reduction scheme employed at each node. However, the gradient tables result in an additional storage cost of~$\mc O(pm_i)$ at each node~$i$. Given a desired accuracy, applicable trade-offs between the convergence benefits of the \textbf{\texttt{GT-VR}} framework versus the simplicity of other methods can be formulated; see also~\cite{SPM2019_Xin} for a detailed review.
\end{rmk}

\begin{rmk}[Other instances of the \textbf{\texttt{GT-VR}} framework]
The discussion in this section enables a suite of methods that combine various gradient tracking and variance reduction techniques leading to different performance and implementation trade-offs towards different problem classes~\cite{SPM2019_Xin,GT_SARAH}. For example, the storage of~\textbf{\texttt{GT-SAGA}} can be avoided in~\textbf{\texttt{GT-SVRG}},  which computes batch gradients periodically, however, at the expense of additional network synchronization~\cite{GTVR_TSP,SPM2019_Xin}. For non-convex finite sum problems, it is recently shown that~\textbf{\texttt{GT-SARAH}}~\cite{GT_SARAH}, that uses variance reduction of~\textbf{\texttt{SARAH}}-type~\cite{SARAH,SPIDER}, achieves an optimal and network-independent performance in a certain regime of practical interest. See also~\cite{D_Get} that uses~\textbf{\texttt{SARAH}}-type variance reduction for non-convex expected risk minimization problems. We also note that a recent work~\cite{PS_MIQ:2020} proposes~\textbf{\texttt{Push-SAGA}} that extends~\textbf{\texttt{GT-SAGA}} to arbitrary directed graphs with the help of push-sum consensus protocol.  
\end{rmk}

\begin{rmk}[Other decentralized variance reduction methods]
Decentralized variance-reduced stochastic gradient methods include \textbf{\texttt{DSA}}~\cite{DSA} that combines \textbf{\texttt{EXTRA}}~\cite{EXTRA} and \textbf{\texttt{SAGA}}~\cite{SAGA}, \textbf{\texttt{diffusion-AVRG}}~\cite{DAVRG} that combines exact diffusion~\cite{Exact_Diffusion} and \textbf{\texttt{AVRG}}~\cite{AVRG}, \textbf{\texttt{DSBA}}~\cite{DSBA} that adds proximal mapping~\cite{point-SAGA} to \textbf{\texttt{DSA}}, \textbf{\texttt{ADFS}}~\cite{ADFS} that applies an accelerated randomized proximal coordinate gradient method~\cite{APCG} to a dual formulation, and \textbf{\texttt{Network-SVRG/SARAH}}~\cite{N_DANE} that implements variance reduction in the \textbf{\texttt{DANE}} framework~\cite{DANE}. We note that when~$M\!\approx\! m$ is very large,~{\GTS} improves upon the convergence rate of these methods in terms of the joint dependence on~$\kappa$ and~$M$, with the exception of \textbf{\texttt{DSBA}} and \textbf{\texttt{ADFS}}, both of which achieves better iteration complexity, however, at the expense of computing proximal mappings. 
\end{rmk}


\section{Numerical Experiments}\label{s_exp}
In this section, we compare the performance of the decentralized first-order methods discussed in this article with the help of linear and logistic regression problems. 

\subsubsection{Decentralized state estimation in sensor networks}
\begin{figure}[!h]
\centering
\subfigure{\includegraphics[width=2.25in]{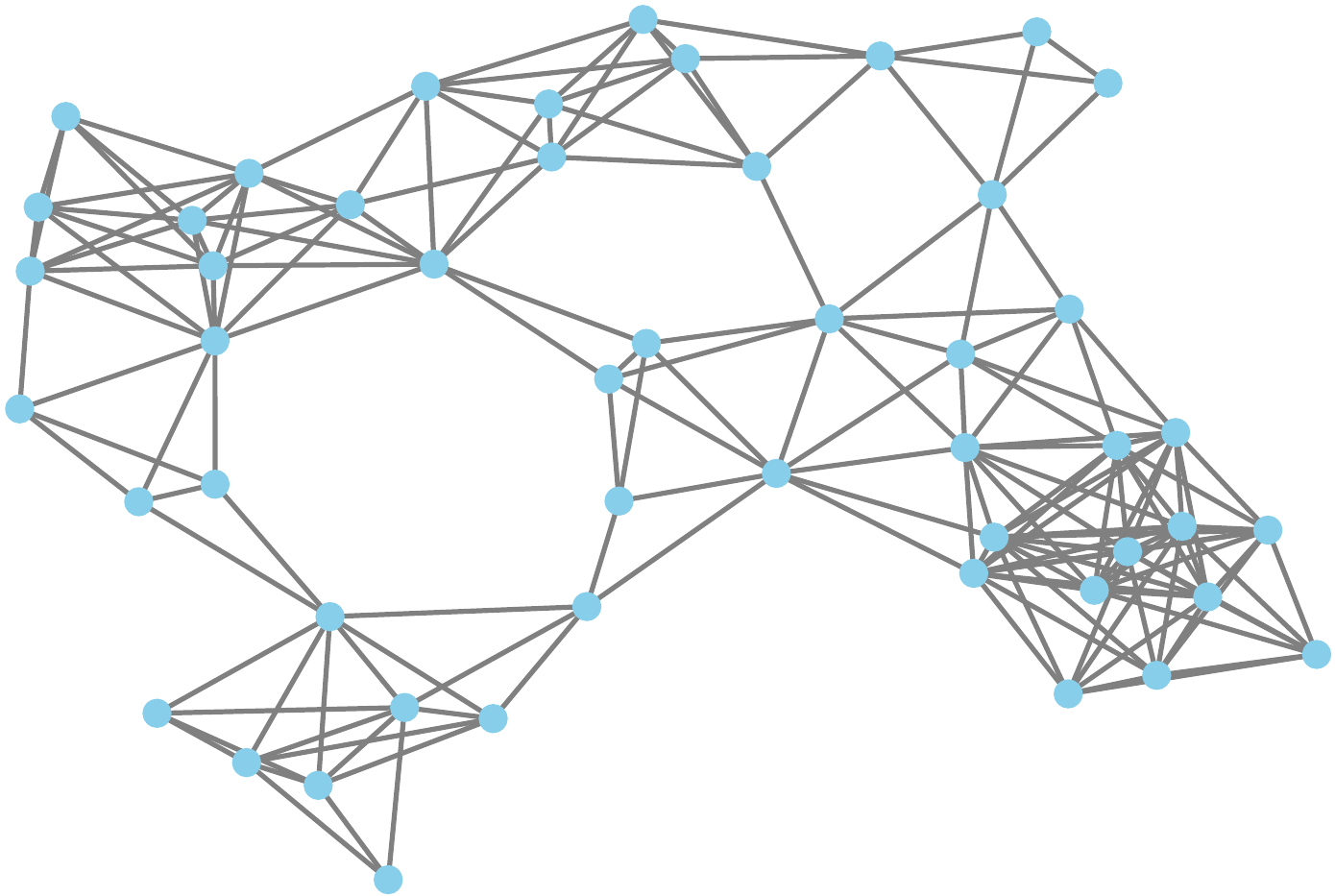}}\\
\subfigure{\includegraphics[width=2.25in]{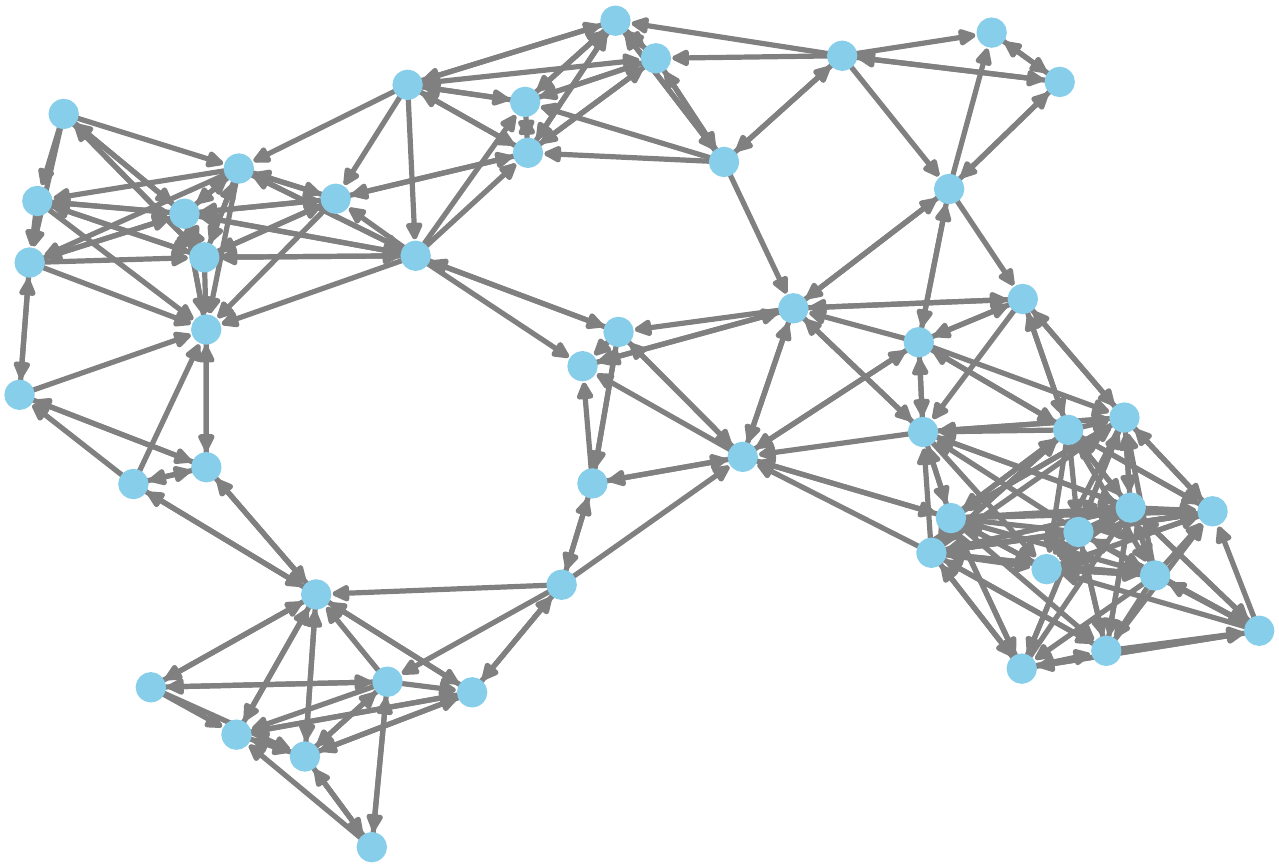}}
\caption{An undirected geometric graph (top) and a directed geometric graph (bottom) generated by nearest neighbor rule.}
\label{UG_geometric}
\end{figure}

Consider a network of sensors deployed to estimate a state vector~${\mb{x}\in\mathbb{R}^p}$, e.g., the location of a target or a physical quantity of interest. Each sensor~$i$ obtains its local measurements~${\mb{y}_i\in\mathbb{R}^{p_i}}$ as~${\mb{y}_i = H_i \mb{x} + \mb{n}_i}$, where~${H_i\in\mathbb{R}^{p_i\times p}}$ is the local sensing matrix and~${\mb{n}_i}$ is some random noise. To recover the state vector, the nodes cooperate to solve the following optimization problem:~${
\min_{\mb{x}\in\mathbb{R}^p}\sum_{i=1}^n\|\mb{y}_i - H_i\mb{x}\|^2.}$
We randomly generate an undirected graph of~$50$ nodes using the nearest neighbor rule, i.e., two nodes are connected if they lie within a communication radius. A directed graph is obtained by making half of the links in the undirected geometric graph one-directional. Two such sample graphs are shown in Fig.~\ref{UG_geometric}. We set the dimension of the state~${\mb x\in\mbb R^p}$ to be~${p = 100}$ and generate the sensing matrices and the state vector from Gaussian distribution with zero-mean and a standard deviation of~$10$. The maximum rank of the sensing matrices does not exceed~$20$, i.e., no node is able to recover the state on its own. 

In Fig.~\ref{LS_UG}, we compare \textbf{\texttt{DGD}}, \textbf{\texttt{GT-DGD}}, and \textbf{\texttt{AB}}/\textbf{\texttt{Push-Pull}} over an undirected graph. Note that doubly stochastic weights are used for \textbf{\texttt{DGD}} and \textbf{\texttt{GT-DGD}} while row and column stochastic weights are used in \textbf{\texttt{AB}}/\textbf{\texttt{Push-Pull}}. The same constant step-size~$\a = 10^{-5}$ is used for all of the algorithms. As discussed before, \textbf{\texttt{DGD}} with a constant step-size has an inexact linear convergence while with a decaying step-size has an exact but sub-linear convergence. Both \textbf{\texttt{GT-DGD}} and \textbf{\texttt{AB}}/\textbf{\texttt{Push-Pull}} converge linearly to the exact solution due to gradient tracking. In Fig.~\ref{LR_all}, we compare Gradient-Push (\textbf{\texttt{GP}}), \textbf{\texttt{ADDOPT}}/\textbf{\texttt{Push-DIGing}} (with only column stochastic weights), \textbf{\texttt{FROST}} (with only row stochastic weights) and \textbf{\texttt{AB}}/\textbf{\texttt{Push-Pull}} over a directed graph. \textbf{\texttt{GP}} being an extension of \textbf{\texttt{DGD}} with push-sum consensus has an inexact linear convergence with a constant step-size while all of the other algorithms are based on gradient tracking and converge to the exact optimal solution linearly. 
\begin{figure}[!h]
\centering
\includegraphics[width=3in]{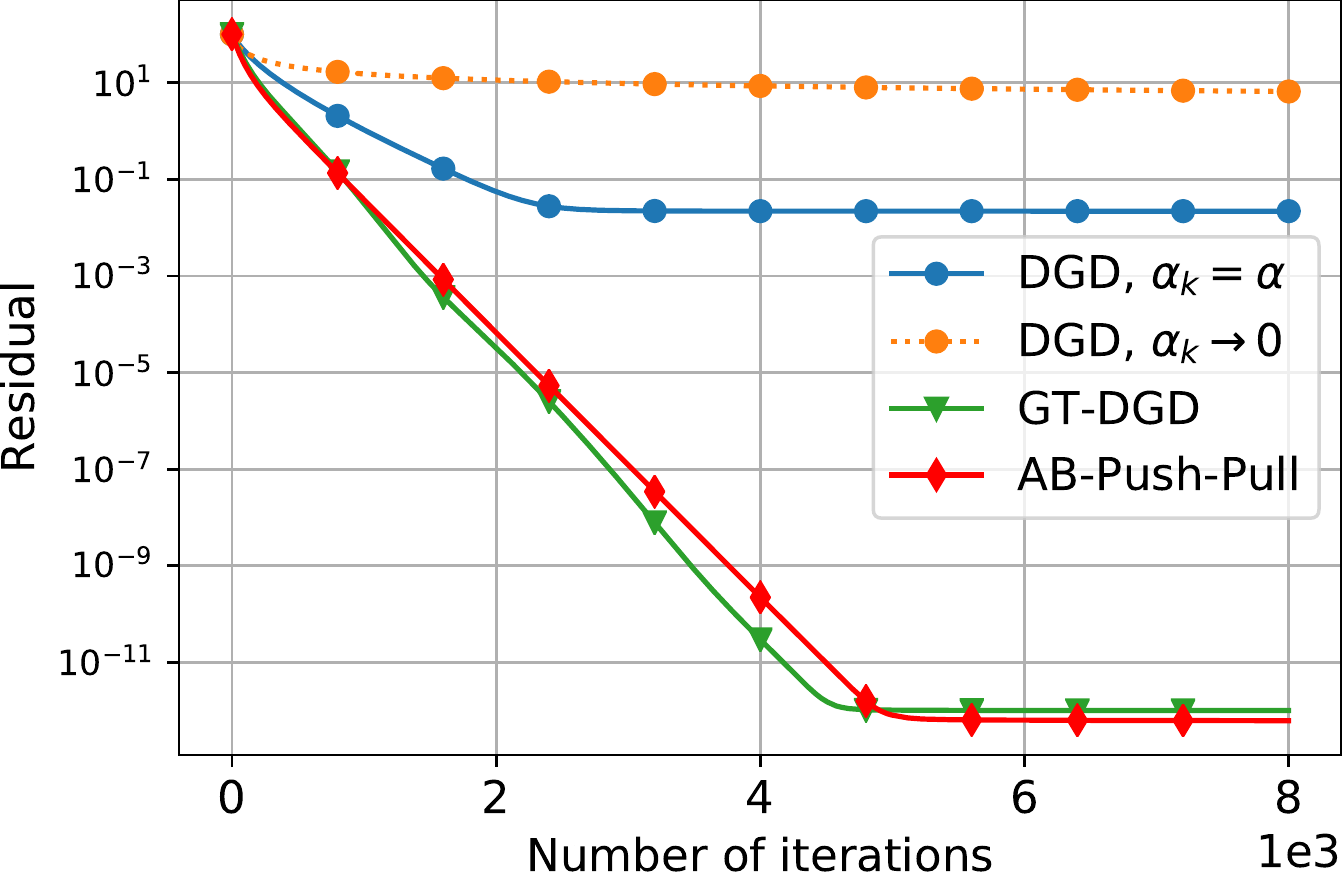}
\caption{Decentralized state estimation over undirected geometric graph.}
\label{LS_UG}
\end{figure}

\begin{figure}[!h]
\centering
\includegraphics[width=3in]{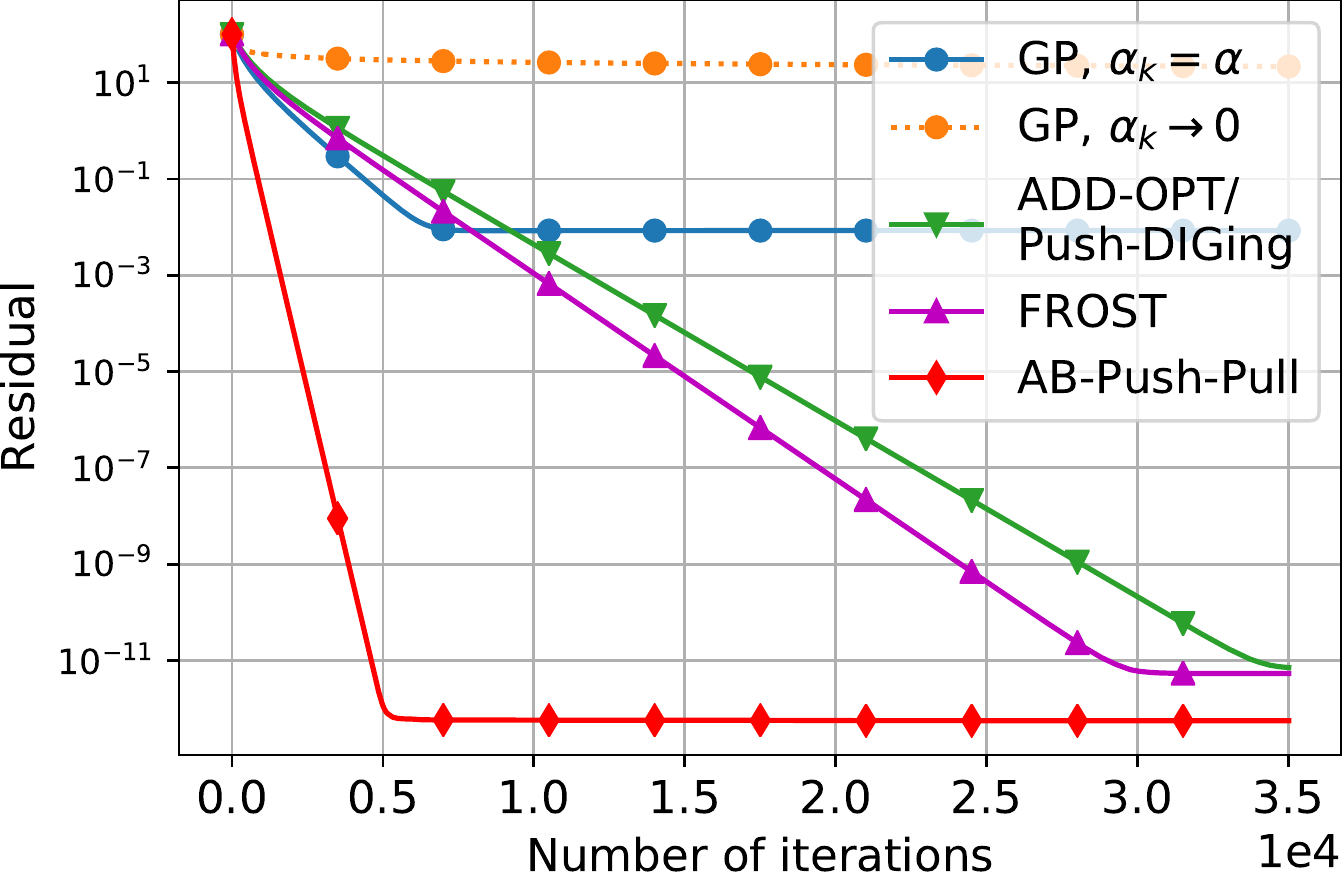}
\caption{Decentralized state estimation over directed geometric graph.}
\label{LS_UG}
\end{figure}

\begin{figure*}[!ht]
\centering
\subfigure{\includegraphics[width=1.75in]{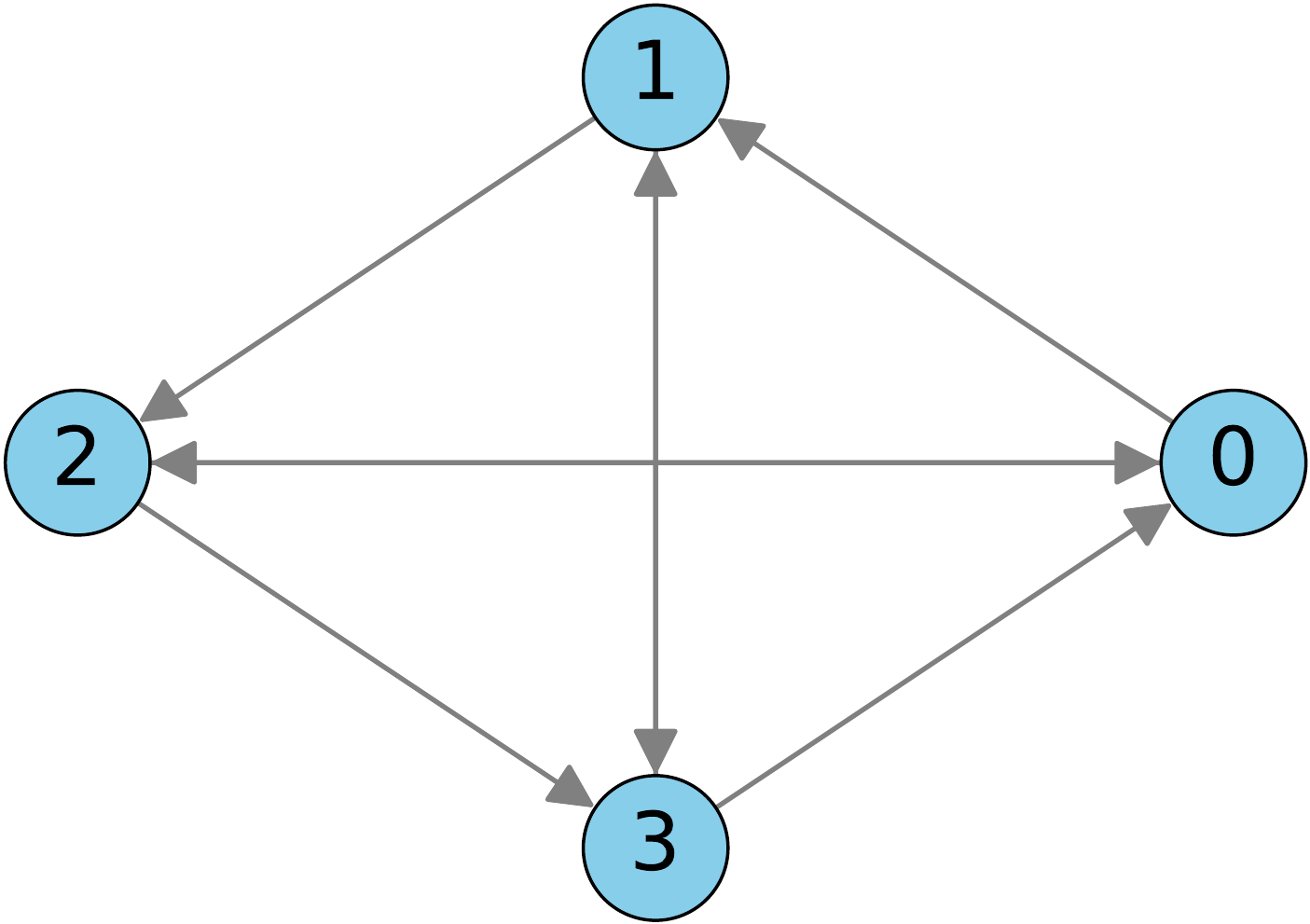}}
\subfigure{\includegraphics[width=1.75in]{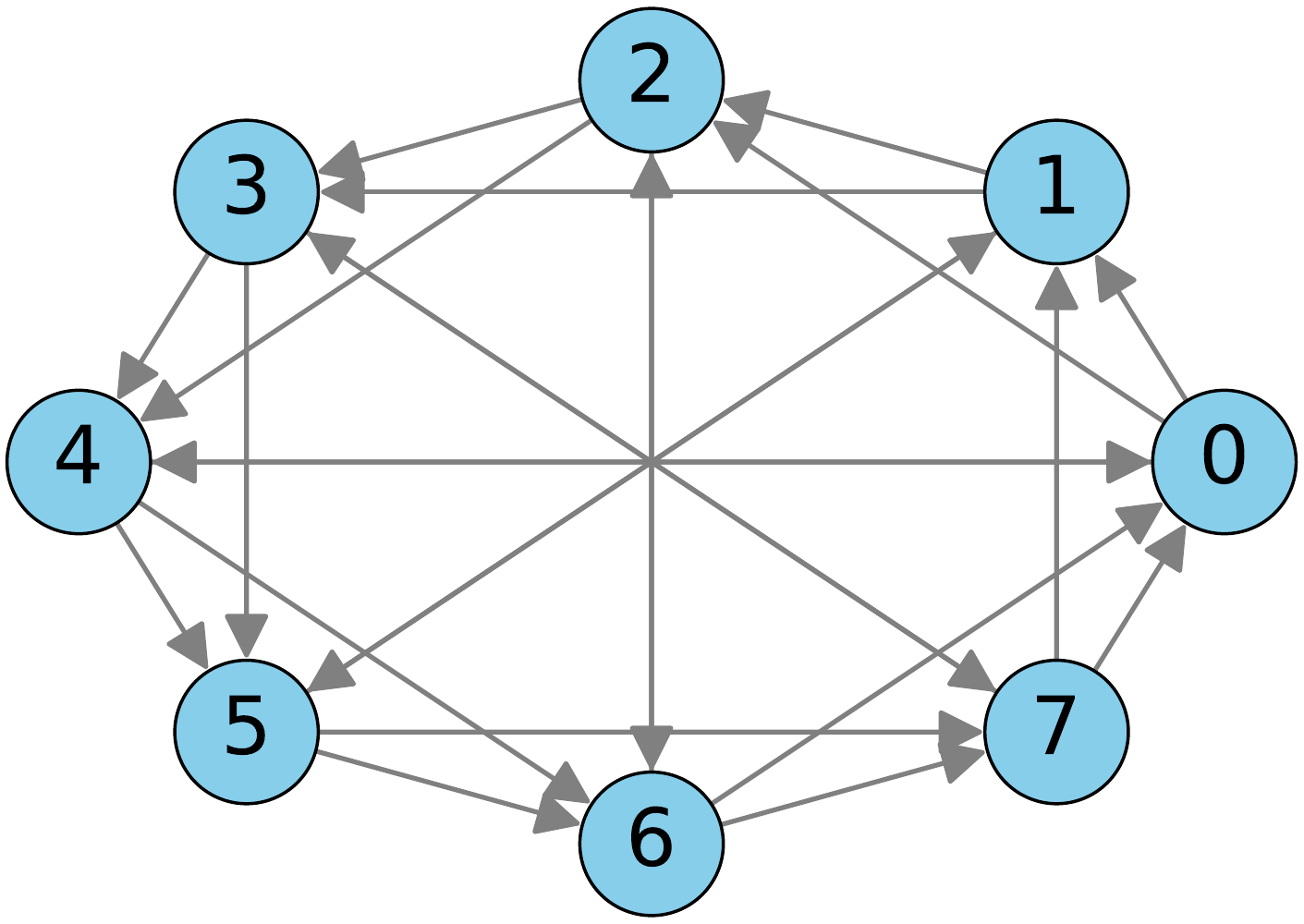}}
\subfigure{\includegraphics[width=1.75in]{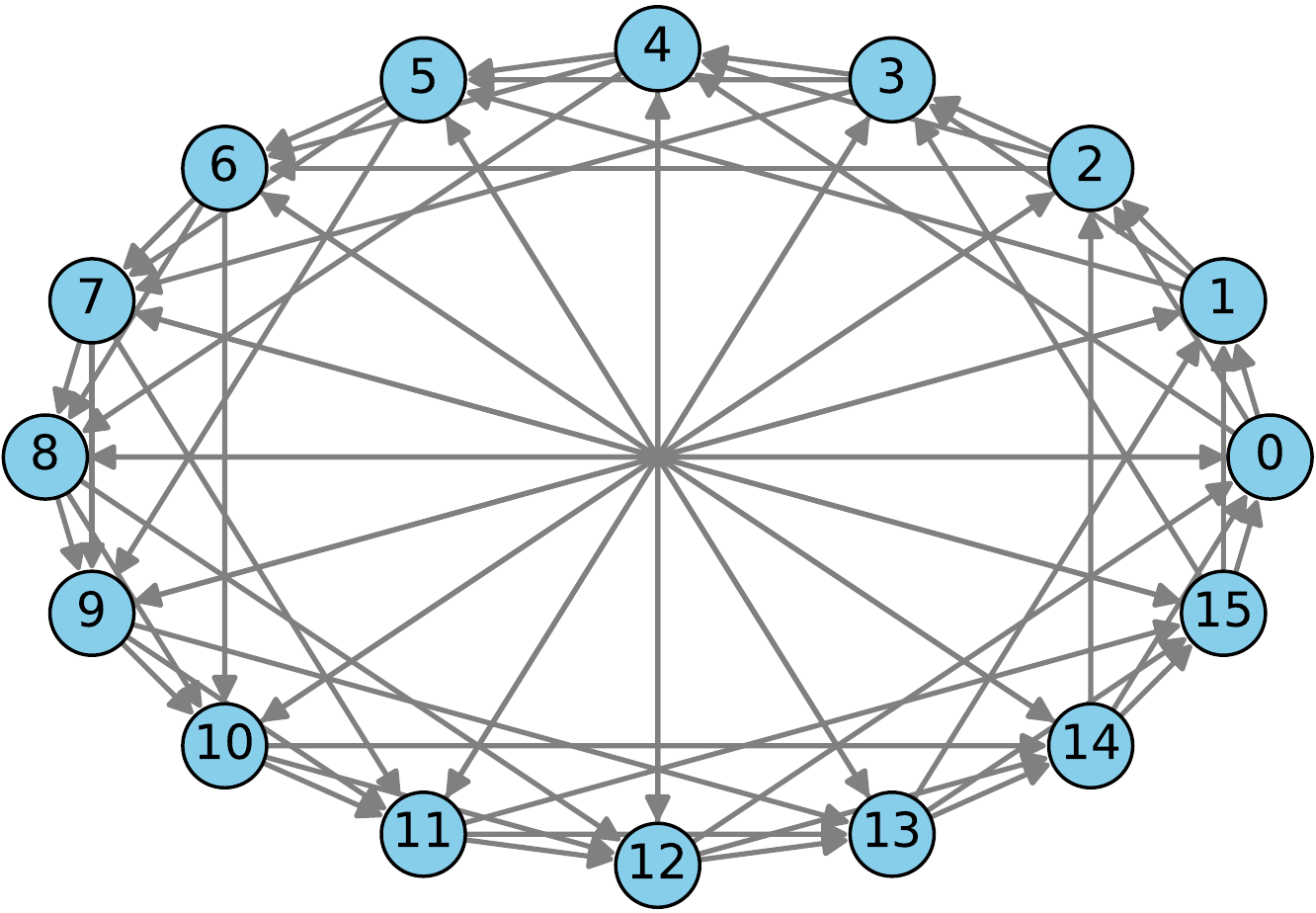}}
\subfigure{\includegraphics[width=1.75in]{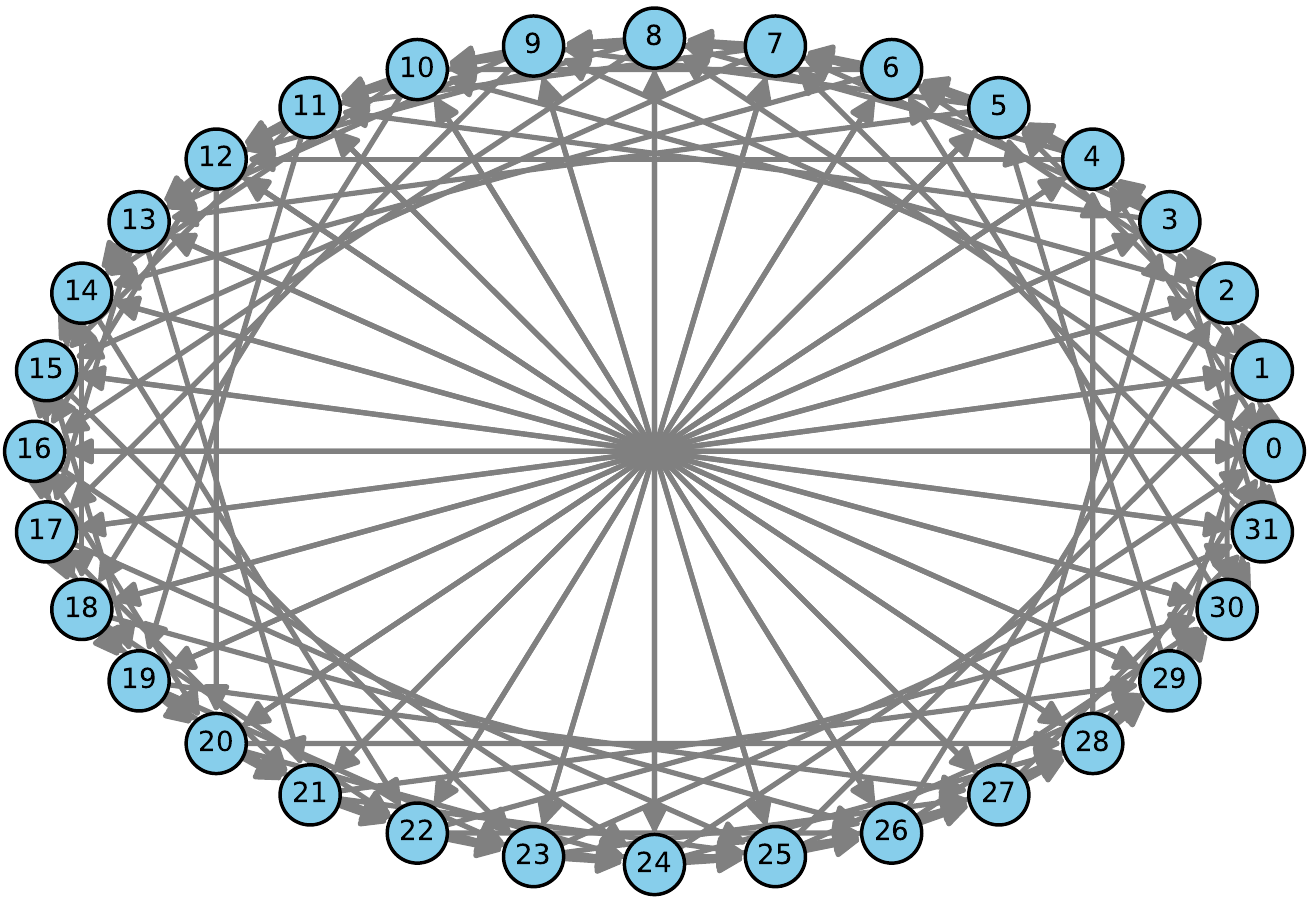}}
\caption{Directed exponential graphs with~$4,8,16$, and~$32$ nodes. These graphs are balanced and thus admit doubly stochastic weights.}
\label{DG_exponential}
\end{figure*}
\subsubsection{Decentralized training over multiple machines}

We next consider decentralized logistic regression over the directed exponential graphs~\cite{SGP_ICML}, shown in Fig.~\ref{DG_exponential}, where the goal is to classify hand-written digits~$3$ and~$8$ from the MNIST dataset~\cite{MNIST}. The class of directed exponential graphs is weight-balanced (therefore admits doubly stochastic weights), sparsely-connected (low-communication cost per node), and possesses a strong algebraic connectivity. The nodes cooperatively solve the following smooth and strongly convex problem:~$\min_{\mb w, b}F(\mb{w},b) = \frac{1}{nm}\sum_{i=1}^{n}\sum_{{j}=1}^{m}f_{i,j}(\mb{w},b)$, with 
\begin{align*}
f_{i,j}(\mb{w},b) &= \ln\left[1+\exp\left\{-(\mb{w}^\top\mb{z}_{{i,j}}+b)y_{i,j}\right\}\right]+\frac{\lambda}{2}\|\mb{w}\|_2^2,
\end{align*}
where~${\mb{z}_{i,j} \in \mathbb{R}^{784}}$ is the feature vector of the~$j$th (digit) image at node~$i$ and and~$y_{i,j}$ is the corresponding binary label. In our setup, a total number of~$11968$ training images are evenly distributed among the nodes. 
We set the regularization parameter~${\lambda = \frac{1}{nm}}$~\cite{SAG}. The global minimum is found by centralized Nesterov gradient descent. We plot the average residual~${\frac{1}{n}\sum_{i=1}^{n}\left(F(x_k^i)-F^*\right)}$ across all nodes for comparison. The hyper-parameters for all algorithms are tuned manually for best performances. 

\begin{figure}[!h]
\centering
\includegraphics[width=3in]{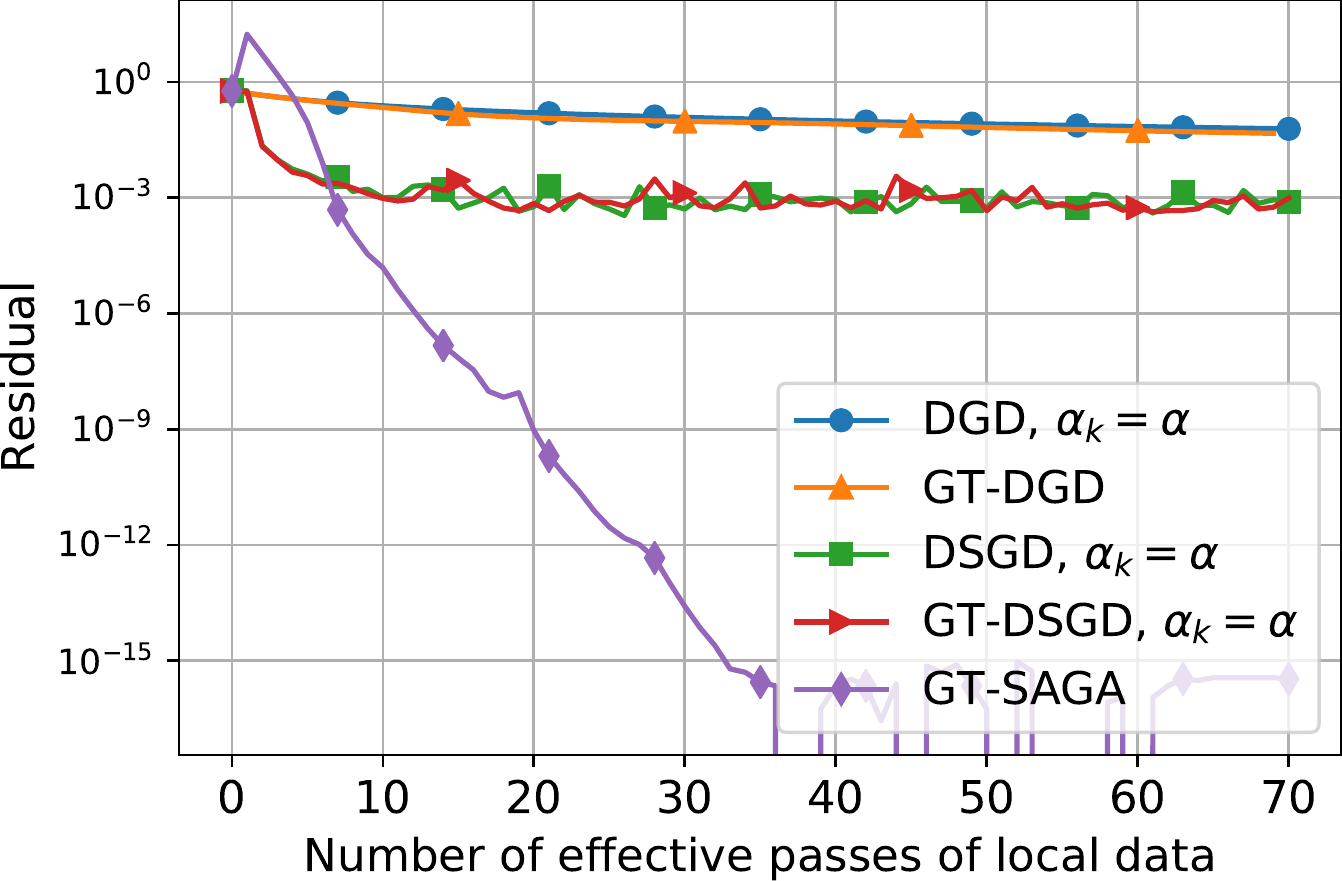}
\caption{Decentralized logistic regression over the directed exponential graph with~$n=8$ nodes.}
\label{LR_all}
\end{figure}

We first show performance comparison over the directed exponential graph with~${n=8}$ nodes (thus there are~${m=1496}$ images per node).
Fig.~\ref{LR_all} compares \textbf{\texttt{DGD}}, \textbf{\texttt{GT-DGD}}, \textbf{\texttt{GT-DSGD}}, \textbf{\texttt{GT-DSGD}}, and \textbf{\texttt{GT-SAGA}}, all with constant step-sizes, where one unit on the horizontal axis (or epoch) represents~$m=1496$ component gradient evaluations at each node, i.e., one effective pass of the local data. It can be observed that in the first few epochs, stochastic gradient methods, \textbf{\texttt{DSGD}} and \textbf{\texttt{GT-DSGD}}, make very fast progress and significantly outperform their deterministic counterparts, \textbf{\texttt{DGD}} and \textbf{\texttt{GT-DGD}}.
Over time, however, stochastic methods drastically slow down and full gradient methods that make steady progress across iterations typically achieve a better performance. Clearly, \textbf{\texttt{GT-SAGA}} that uses both gradient tracking and variance-reduction exhibits consistent fast linear convergence to the global minimum and outperforms all other methods.  

\begin{figure}[!h]
\centering
\includegraphics[width=3in]{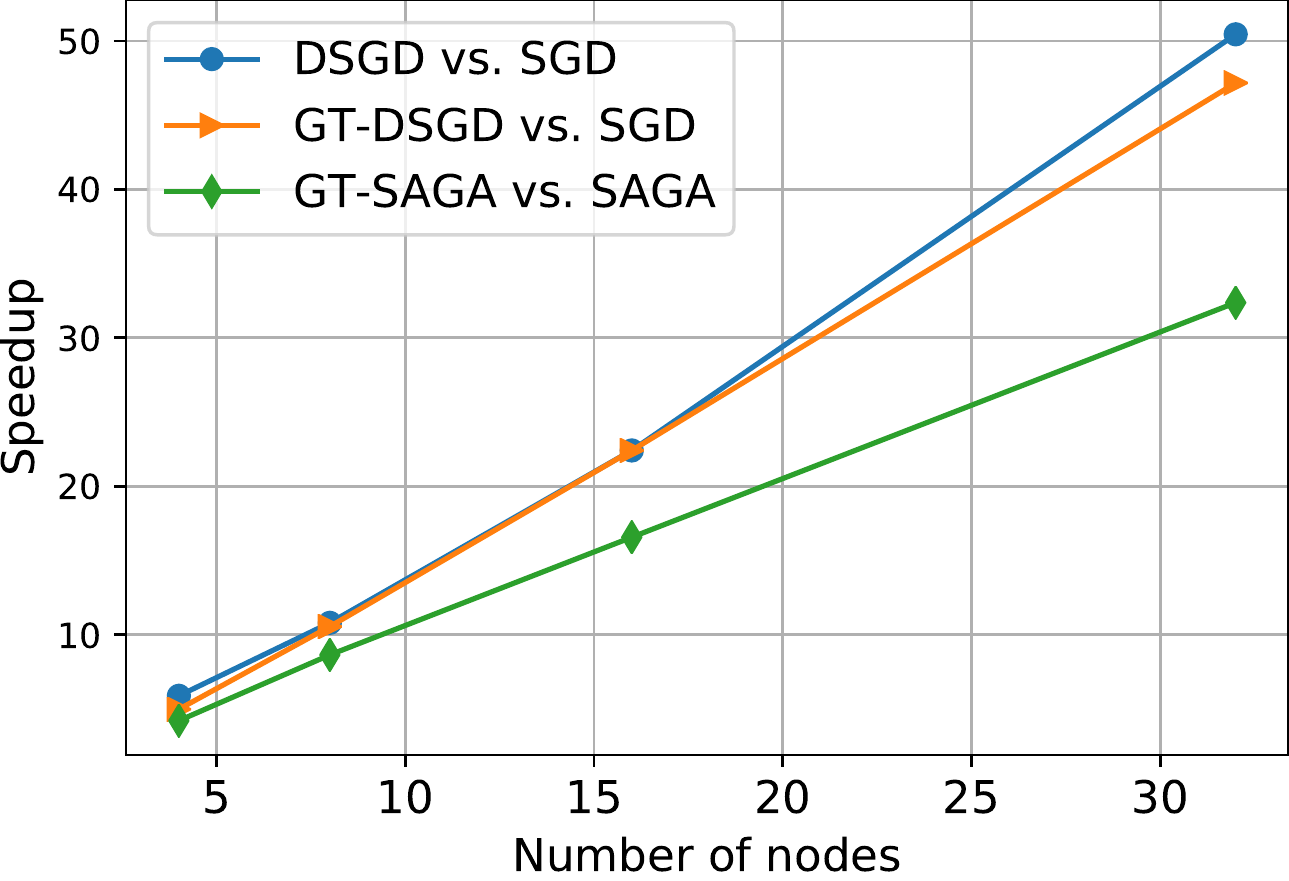}
\caption{Linear speedup: \textbf{\texttt{DSGD}}  and \textbf{\texttt{GT-DSGD}} vs. \textbf{\texttt{SGD}} to achieve an accuracy of~$10^{-3}$; \textbf{\texttt{GT-SAGA}} vs. \textbf{\texttt{SAGA}} to achieve an accuracy of~$10^{-15}$.}
\label{d_speedup}
\end{figure}

Finally, we study the speedup achieved by the decentralized stochastic methods over their centralized counterparts in Fig.~\ref{d_speedup}, where we compare \textbf{\texttt{DSGD}} and \textbf{\texttt{GT-DSGD}} with centralized \textbf{\texttt{SGD}}, and \textbf{\texttt{GT-SAGA}} with centralized \textbf{\texttt{SAGA}}. Each iteration of centralized \textbf{\texttt{SGD}} and \textbf{\texttt{SAGA}} evaluates one component gradient while each iteration of \textbf{\texttt{DSGD}}, \textbf{\texttt{GT-DSGD}} and \textbf{\texttt{GT-SAGA}} evaluates~$n$ component gradients in parallel, i.e., one at each node. The speedup can be described as the the ratio of the number of iterations taken by the centralized method over the corresponding decentralized method to achieve the same accuracy ($10^{-3}$ for comparisons with \textbf{\texttt{SGD}} and~$10^{-15}$ for comparison with \textbf{\texttt{SAGA}}). We conduct the experiment over the directed exponential graphs with~$4,8,6$, and~$32$ nodes shown in Fig.~\ref{DG_exponential}.
It can be observed that a linear speedup is achieved for all three decentralized approaches. In other words, the decentralized methods are  approximately~$n$ times faster than their centralized counterparts and thus are particularly favorable when data can be processed in parallel.

\section{Conclusions}\label{s_conc}
In this article, we study decentralized first-order methods with full and stochastic gradients over undirected and directed graphs. We show that most of the existing work on decentralized methods based on gradient tracking can be cast in a unifying framework provided by the~$\AB$/\textbf{\texttt{Push-Pull}} algorithm. On the stochastic front, we discuss how gradient tracking is unable to achieve exact linear convergence, which can be recovered with the addition of variance-reduction. We provide numerical results to illustrate the convergence  behavior and performance of the corresponding methods.

\bibliographystyle{Aux/IEEEbib}
\bibliography{KHAN_allPUBS,sample}

\begin{IEEEbiography}[{\includegraphics[width=1in,height=1.2in,clip,keepaspectratio]{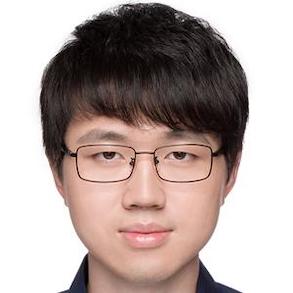}}]{Ran Xin} received his B.S. degree in Mathematics and Applied Mathematics from Xiamen University, China, in 2016, and M.S. degree in Electrical and Computer Engineering from Tufts University in 2018. Currently, he is a Ph.D. candidate in the Electrical and Computer Engineering Department at Carnegie Mellon University. His research interests include convex and nonconvex optimization, stochastic approximation and machine learning.
\end{IEEEbiography}

\vspace{-0.6cm}
\begin{IEEEbiography}[{\includegraphics[width=1in,height=1.2in,clip,keepaspectratio]{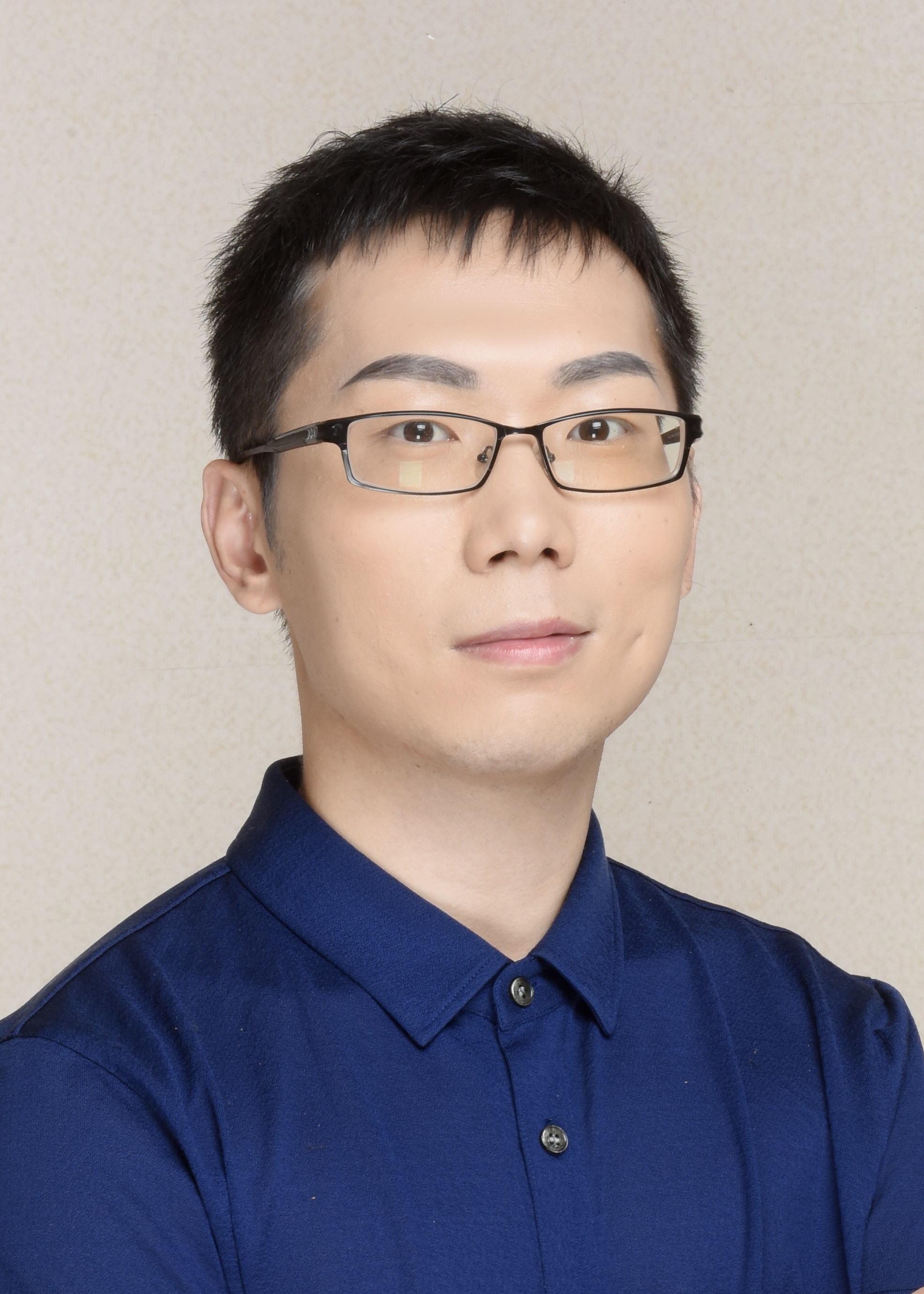}}]{Shi Pu} is currently an assistant professor in the School of Data Science, The Chinese University of Hong Kong, Shenzhen, China. He received a B.S. Degree in Engineering Mechanics from Peking University, in 2012, and a Ph.D. Degree in Systems Engineering from the University of Virginia, in 2016. He was a postdoctoral associate at the University of Florida, from 2016 to 2017, a postdoctoral scholar at Arizona State University, from 2017 to 2018, and a postdoctoral associate at Boston University, from 2018 to 2019. His research interests include distributed optimization, network science, machine learning, and game theory.
\end{IEEEbiography}

\vspace{-0.6cm}
\begin{IEEEbiography}[{\includegraphics[width=1in,height=1.2in,clip,keepaspectratio]{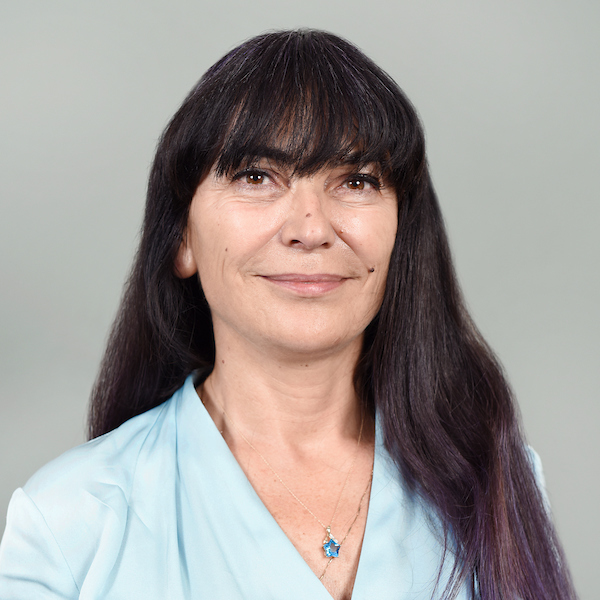}}]{Angelia Nedi\'{c}} has a Ph.D. from Moscow State University, Moscow, Russia, in Computational Mathematics and Mathematical Physics (1994), and a Ph.D. from Massachusetts Institute of Technology, Cambridge, USA in Electrical and Computer Science Engineering (2002). She has worked as a senior engineer in BAE Systems North America, Advanced Information Technology Division at Burlington, MA. Currently, she is a faculty member of the school of Electrical, Computer and Energy Engineering at Arizona State University at Tempe. Prior to joining Arizona State University, she has been a Willard Scholar faculty member at the University of Illinois at Urbana-Champaign. She is a recipient (jointly with her co-authors) of the Best Paper Awards at the Winter Simulation Conference 2013 and at the International Symposium on Modeling and Optimization in Mobile, Ad Hoc and Wireless Networks (WiOpt) 2015.  Her general research interest is in optimization, large scale complex systems dynamics, variational inequalities and games.
\end{IEEEbiography}

\vspace{-0.6cm}
\begin{IEEEbiography}[{\includegraphics[width=1in,height=1.2in,clip,keepaspectratio]{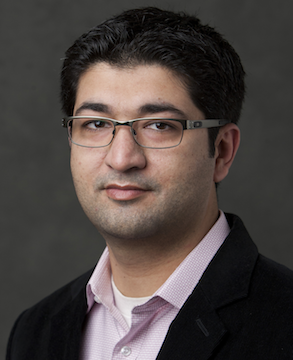}}]{Usman A. Khan} is an Associate Professor of Electrical and Computer Engineering (ECE) at Tufts University, Medford, MA, USA. His research interests include statistical signal processing, network science, and decentralized optimization over autonomous multiagent systems. Recognition of his work includes the prestigious National Science Foundation (NSF) Career award, an IEEE journal cover, three best student paper awards in IEEE conferences, and several news articles including two in \textit{IEEE Spectrum}. He received his B.S. degree in 2002 from University of Engineering and Technology, Pakistan, M.S. degree in 2004 from University of Wisconsin-Madison, USA, and Ph.D. degree in 2009 from Carnegie Mellon University, USA, all in ECE. Dr. Khan is an \textit{IEEE Senior Member} and has been an elected full member of the \textit{Sensor Array and Multichannel Technical Committee} with the \textit{IEEE Signal Processing Society} since 2019 where he was an Associate member from 2010 to 2019. He was an elected full member of the \textit{IEEE Big Data special interest group} from 2017 to 2019. He was an Editor of the \textit{IEEE Transactions on Smart Grid} from 2014 to 2017 and is currently an Associate Editor of the \textit{IEEE Control System Letters}, \textit{IEEE Transactions Signal and Information Processing over Networks}, and \textit{IEEE Open Journal of Signal Processing}. He is the Lead Guest Editor for the \textit{Proceedings of the IEEE Special Issue on Optimization for Data-driven Learning and Control} and a Guest Associate Editor for \textit{IEEE Control System Letters Special Issue on Learning and Control} both to appear in 2020. He is the Technical Area Chair for the Networks track in \textit{IEEE 2020 Asilomar Conference on Signals Systems and Computers}. 
\end{IEEEbiography}

\end{document}